\documentclass[10pt,twocolumn,letterpaper]{article}
\usepackage{amsmath,amsthm}
\usepackage[numbers,sort&compress]{natbib}
\usepackage{thunlp_cvpr}
\usepackage{latexsym}

\usepackage{tabularx}

\definecolor{mentalblue}{RGB}{235,243,250}
\usepackage{inconsolata}
\usepackage{mathtools}

\theoremstyle{plain}

\theoremstyle{definition}

\theoremstyle{remark}

\definecolor{def}{RGB}{119, 228, 200}
\definecolor{thm}{RGB}{69, 53, 193}
\definecolor{boxred}{RGB}{190, 30, 45} 
\definecolor{lightred}{RGB}{255, 245, 245}

\newtcolorbox{thmbox}[1][]{colback=thm!5!white,colframe=thm!60!black,boxsep=-4pt,grow to left by=4pt,left=10pt,grow to right by=4pt,right=10pt,top=10pt,bottom=10pt,#1}
\newtcolorbox{redbox}[1][]{colback=boxred!5!white,colframe=boxred!90!black,boxsep=-4pt,grow to left by=4pt,left=10pt,grow to right by=4pt,right=10pt,top=10pt,bottom=10pt,#1}
\newtcolorbox{defbox}[1][]{colback=def!5!white,colframe=def!60!black,boxsep=-4pt,grow to left by=4pt,left=10pt,grow to right by=4pt,right=10pt,top=10pt,bottom=10pt,#1}

\usepackage{listings}
\lstdefinestyle{promptstyle}{
  basicstyle=\ttfamily\small,
  breaklines=true,
  frame=single,
  backgroundcolor=\color{white},
  keywordstyle=\color{blue},
  commentstyle=\color{gray},
  captionpos=b
}

\usepackage{nicematrix}
\usepackage{makecell}

\newcommand{\gain}[1]{\textcolor{green!60!black}{#1}}

\newcommand{\twolinescore}[2]{\makecell[c]{#1\\[-0.4ex]
        {\fontsize{4.5pt}{4.8pt}\selectfont\mdseries #2}}}

\usepackage{xspace,colortbl}

\newcommand{\benchname}{\textbf{BASIS}\xspace}
\usepackage{url}
\usepackage[utf8]{inputenc}

\title{BASIS: Evaluating Belief-Aware Solutions in Interactive Settings beyond Instructions}

\author{
  Minyuan Ruan\textsuperscript{1*}\quad
  Ziyue Wang\textsuperscript{1*}\quad
  Kaiming Liu\textsuperscript{1,3}\quad
  Yunghwei Lai\textsuperscript{1}\quad
  Peng Li\textsuperscript{2,$\dagger$}\quad
  Yang Liu\textsuperscript{1,2,$\dagger$}\\
  \thunlpauthoraffildivider
}
\thunlpaffiliation{
  \textsuperscript{1}Dept. of Comp. Sci. \& Tech., Institute for AI, Tsinghua University, Beijing, China\\
  \textsuperscript{2}Institute for AI Industry Research (AIR), Tsinghua University, Beijing, China\\
  \textsuperscript{3}College of AI, Tsinghua University, Beijing, China
}
\date{}
\thunlplogooff
\setthunlparticlemeta{RESEARCH ARTICLE}

\begin{document}

\makefrontmatter{
Large language model (LLM) agents are increasingly capable of carrying out complex, long-horizon tasks, yet effective assistance requires more than faithfully executing user instructions. Users' requests reflect their current understanding of the problem, which may be incomplete, outdated, or incorrect. When this understanding diverges from the actual task state, literal instruction following can fail to accomplish the user's underlying goal. We introduce \benchname (\emph{\textbf{B}elief-\textbf{A}ware \textbf{S}olutions in \textbf{I}nteractive \textbf{S}ettings}), a benchmark that evaluates whether agents can infer the mental states underlying user requests, recognize belief-reality mismatches, and use this understanding to develop appropriate solutions. Rather than assessing mental-state inference solely through explicit questions, \benchname evaluates both the utility of such inferences and the quality of downstream solutions in user-agent interactions. Evaluation results show that current models often fail to account for users' mental states when generating solutions, and that solution quality is consistently higher when such information is explicitly elicited and made available during problem solving. We further investigate the practical value of mental-state consideration through comparisons with generic Chain-of-Thought prompting, controlled interventions on mental-state information, targeted training, and multi-agent collaboration experiments under partial observability. Together, these studies highlight the importance of not merely inferring others' mental states, but also using those understandings to provide grounded, goal-directed assistance.
\footnote{Code and data are available at: \url{https://github.com/THUNLP-MT/BASIS}.}
}

\begingroup
\renewcommand{\thefootnote}{\fnsymbol{footnote}}
\footnotetext[1]{Equal contribution.}
\footnotetext[2]{Corresponding author: pengli09@gmail.com,\\ liuyang2011@tsinghua.edu.cn.}
\endgroup

\section{Introduction}

Large language model (LLM) agents are becoming increasingly capable of executing complex instructions over extended horizons~\citep{openai2026gpt6astra,anthropic2026fablemythos}. Yet effective interaction requires more than faithfully carrying out an instruction as given. 
Agents must also reason about the mental states that give rise to users' requests, like what they believe, know, intend, and prefer, and use such reasoning to decide how to act. 

\begin{figure*}[t]
    \centering
    \includegraphics[width=\textwidth]{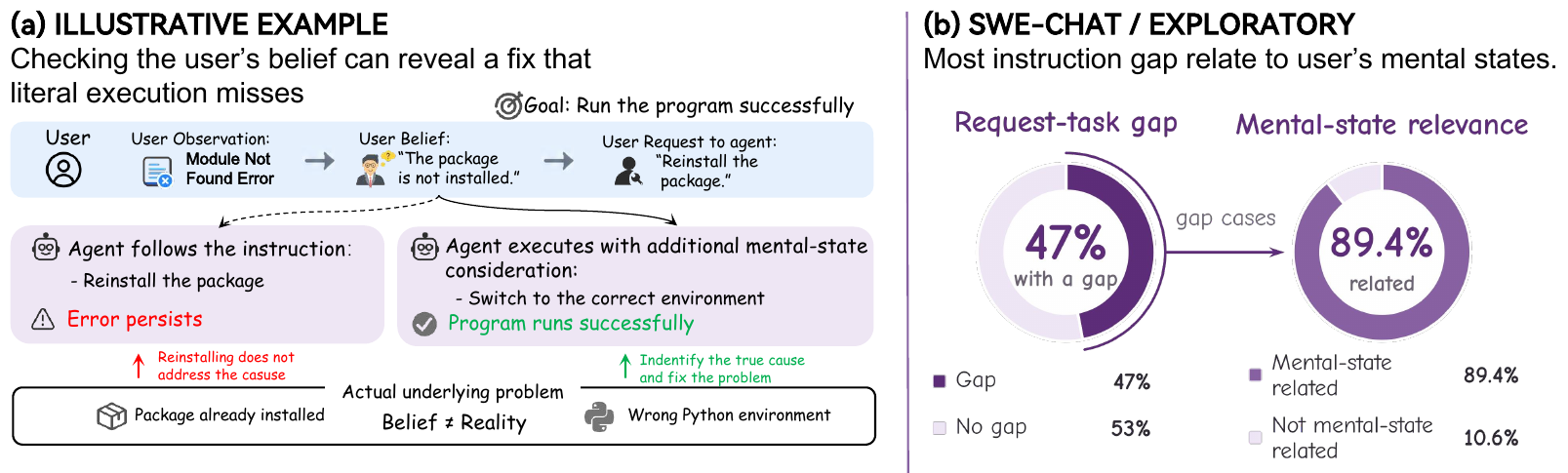}
\caption{
    \textbf{(a)} An illustrative case shows a user interpreting a \texttt{ModuleNotFoundError} as a missing package and requesting reinstallation. Literal execution repeats the installation and leaves the error unresolved, whereas mental-state consideration helps the agent to check the active Python environment and use the one where the package is already installed.
    \textbf{(b)} An exploratory annotation study of SWE-Chat identifies instruction gaps, where mismatches exist between users' explicit requests and what is needed to achieve their underlying goals, in 47\% of interactions; among these gap cases, 89.4\% show direct or strong contextual evidence related to users' mental states.
}
\label{fig:teaser}
\end{figure*}

This is particularly important when a user request reflects an incomplete or inaccurate understanding of the underlying task state. Like the case illustrated in Figure~\ref{fig:teaser}(a), the user's instruction is not a direct description of the environment, but an action generated from his current mental model of the problem.
It may diverge from the actual task state due to access to only partial observations, outdated beliefs about the environment, or assumptions that are not supported by the current state.
Consequently, a gap can arise between what the user asks the agent to do and what would actually resolve the underlying problem.
When such a gap is consequential, faithfully executing the request can reinforce the user's mistaken premise rather than help achieve their goal.

Our analysis of real-world interactions from SWE-Chat~\citep{baumann2026swechatcodingagentinteractions} suggests that these gaps are common rather than exceptional.
As shown in Figure~\ref{fig:teaser}(b), approximately 47\% of user instructions are incomplete or incorrect with respect to the underlying task.
Among these gap cases, many of them are associated with factors in the user's mental state including limited observation and personal preferences or assumptions.
Despite their prevalence, agents rarely respond by incorporating these factors into their behavior.
Among instances with flawed instructions, agents recognize the discrepancy and adapt their actions accordingly in only 19.1\% of cases. Instead, they literally follow the instruction or restrict their response to clarification in most cases.
Details of the SWE-Chat annotation study are provided in Appendix~\ref{app:swechat_annotation}.

A growing body of work examines whether LLMs and agents can infer users' mental states~\citep{ma-etal-2023-towards-holistic,chen2025theorymindlargelanguage}. However, these studies primarily evaluate whether models can recover what users believe, know, or intend, while paying less attention to whether such inferences are actually used to guide subsequent behavior~\citep{riemer2025positiontheorymindbenchmarks}. In interactive settings, an agent may correctly infer a user's mental state when explicitly queried, yet still treat the user's instruction as a complete task specification and act on it directly. Thus, successful mental-state inference does not necessarily translate into belief-aware action.

Motivated by this gap, we introduce \benchname (\emph{\textbf{B}elief-\textbf{A}ware \textbf{S}olutions in \textbf{I}nteractive \textbf{S}ettings}), a benchmark that evaluates whether agents can infer user mental states and exploit these inferences during interactive problem solving. We focus on cases where users form requests based on beliefs that are inconsistent with the actual state of the environment. In such settings, successful assistance requires more than identifying what the user believes or detecting a mistaken assumption; the agent must also use this understanding to adapt its response and develop a solution that advances the user's underlying goal. Rather than evaluating mental-state inference solely through explicit questions, \benchname examines whether agents recognize belief-reality mismatches and account for them in their subsequent solutions.

Our experiments show that models perform better when explicitly prompted to infer the user's mental state and apply that inference compared to asking for mental-state inferences and solutions in separate calls. It highlights the importance of connecting mental-state inference to downstream problem solving. We further compare explicit mental-state consideration with generic Chain-of-Thought prompting and conduct controlled interventions to examine the role of task-aligned mental-state information in solution quality. We also investigate whether targeted training can improve agents' use of mental-state inferences. Finally, we extend the study to multi-agent collaboration through experiments like AgentWebBench~\citep{zhong2026agentwebbench} in which agents receive only partial information, examining the relevance of mental-state consideration beyond user assistance.

Our contributions are as follows:

\begin{itemize}
    \item We identify and formalize belief-reality mismatch in
    user-agent interaction, where instructions grounded in incomplete, outdated, or incorrect beliefs may fail to advance the user's underlying goal even when faithfully executed.

    \item We introduce \benchname, a benchmark that evaluates mental-state inference and downstream solution quality in belief-reality mismatch scenarios. Evaluation results show that explicitly using inferred mental states generally improves task resolution.

    \item We investigate the practical value and improvement of mental-state consideration through generic reasoning baselines, controlled interventions, and targeted training, and extend the evaluation to multi-agent collaboration under partial observability.
\end{itemize}

\section{Related Works}
\subsection{Mental state inference}
In psychology, the capacity to attribute mental states to others refers to the term Theory of Mind (ToM)~\citep{premack1978does,apperly2010mindreaders}. This capacity supports a broad range of social functions, including communication, empathy and conflict resolution~\citep{Jarvis_Wong_Weightman_Keage_2024}. Early studies probed ToM in language models through adaptations of classical psychological tasks, particularly false-belief tests~\citep{bubeck2023sparksartificialgeneralintelligence,vanduijn2023theorymindlargelanguage,street2024llm}. Subsequent benchmarks expanded this evaluation to more diverse social contexts, mental-state types, and orders of recursive reasoning, including ToMBench~\citep{chen2024tombench}, BigToM~\citep{gandhi2023understanding}, Hi-ToM~\citep{he2023hitombenchmarkevaluatinghigherorder} among others~\citep{soubki-etal-2024-views,xu2024opentom}. These benchmarks primarily assess whether a model can infer and report another character's mental state from a provided context. However, success on them does not by itself demonstrate that the inferred state is used to guide downstream behavior~\citep{riemer2025positiontheorymindbenchmarks}.

\subsection{From Mental-State Inference to Functional ToM}
While more and more studies investigate whether LLMs can infer others' mental state in more realistic settings~\citep{kim2023fantom,xiao2025dynamictheorymindevaluating,chan2024negotiationtom}, and multimodal scenarios~\citep{jin2024mmtom,shi2025muma}, most still focus on whether models can infer or report others' mental states. Functional ToM, the ability that models use such inferences to guide downstream prediction and action, remains comparatively underexplored. Efforts such as SimpleToM ~\citep{gu2026simpletomexposinggapexplicit} examine whether inferred beliefs support behavioral prediction and judgment, while EnactToM~\citep{juneja2026enacttomevolvingbenchmarkfunctional} extends this evaluation to embodied multi-agent task execution. Our work instead studies whether agents can account for the user's mental state that may diverge from the actual environment state, and respond accordingly rather than executing it blindly.

\subsection{user-agent Interaction}
In parallel, an increasing number of benchmarks study user-agent interaction, where agents must handle under-specified, ambiguous, or incrementally revealed instructions~\citep{yao2024taubenchbenchmarktoolagentuserinteraction,qian2025userbench,barres2025tau,wu2026swetogetherevaluatingcodingagents}. Existing work primarily addresses such uncertainty by asking clarifying questions, eliciting preferences, or inferring latent user intent~\citep{vijayvargiya2025interactive,sun2025training,zhou2025tom,kim2026discoverllmexecutingintentsdiscovering}. However, it often implicitly treats the user as a reliable and stable source of task-relevant information. In practice, users reason from subjective and partial perspectives: their beliefs about the environment may be incorrect or outdated, and their preferences may change throughout an interaction. Our work focuses on this overlooked setting, requiring agents to infer the user's mental state and recognize when a request is grounded in assumptions that diverge from the actual environment state.

\section{Task Formalization}
\label{sec:task_definition}

\paragraph{User requests and belief-reality mismatch.}
The user interprets the partial observation $o$ through a
task-relevant profile $\mathcal{P}$, which may include background,
expertise, preferences, and other characteristics relevant to the task.
This interpretation gives rise to a latent belief $b$ about the
underlying state, which in turn shapes the explicit request $g$ issued
to the agent. We define a \emph{belief-reality mismatch} as a setting in
which a task-relevant assumption encoded in $b$ is inconsistent with
the underlying state $s^*$.

Figure~\ref{fig:teaser} illustrates this process with a simple Python environment example. The underlying state $s^*$ is that the required package is already installed, but the program is running in a different Python environment. The user observes only a \texttt{ModuleNotFoundError} ($o$). Given limited awareness of the active environment ($\mathcal{P}$), the user forms the belief that the package is missing ($b$), leading to the request to reinstall it ($g$). In this case, faithfully executing $g$ is consistent with the user's belief but does not resolve the error. Effective assistance therefore requires the agent to infer the perspective underlying the request, recognize its mismatch with the task state, and switch to the correct environment to achieve the user's goal.

\paragraph{Evaluation task.}
We evaluate whether agents can infer the mental states underlying users'
requests and use these inferences to generate appropriate solutions.
Given a user observation $o$, request $g$, and the first $k$ turns of
interaction history $\tau_{\leq k}$, the model receives
\begin{equation}
    \label{eq:input}
    x=(o,g,\tau_{\leq k}),
\end{equation}
where $k\in\{0,5,10\}$ and $\tau_{\leq 0}=\varnothing$. Reference
annotations of the latent belief $b$, user profile $\mathcal{P}$, and
underlying state $s^*$ are withheld. Conceptually, belief inference can be viewed as posterior inference:
\begin{equation}
    \label{eq:posterior}
    P(b\mid o,g,\tau)
    \propto
    P(\tau\mid o,g,b)\,P(b\mid o,g),
\end{equation}
where the likelihood measures how well a candidate belief explains the
observed interaction, and the prior captures plausible beliefs given the
user's observation and request. We use this formulation as a conceptual
description of mental-state inference.

We assess three targets: the inferred user belief $\widehat{b}$ (\textit{Belief}), the inferred task-relevant profile $\widehat{\mathcal{P}}$ (\textit{Profile}), and the proposed solution $a$ (\textit{Solution}). The solution is evaluated on whether it addresses the actual problem and appropriately handles the relevant mismatch. We compare two evaluation settings.

\noindent\textbf{(1) Separate evaluation.}
Given $x$, the model is asked to infer $\widehat{b}$, infer $\widehat{\mathcal{P}}$, and provide a solution $a_{\mathrm{sep}}$ in independent calls. The solution call asks the model to assist the user without explicitly requesting mental-state inference, and does not receive the outputs of the belief or profile inference calls.

\noindent\textbf{(2) Inference-guided evaluation.}
Given the same input $x$, the model is explicitly instructed to infer $\widehat{b}$ and $\widehat{\mathcal{P}}$ and use these inferences to formulate a solution consistent with the inferred belief and profile. The inferred states and the solution $a_{\mathrm{guide}}$ are generated together in a single call. Comparing solution quality across the two settings examines the benefit of explicitly connecting mental-state inference to downstream problem solving.

\section{benchmark}

\begin{figure*}
    \centering
    \includegraphics[width=1\linewidth]{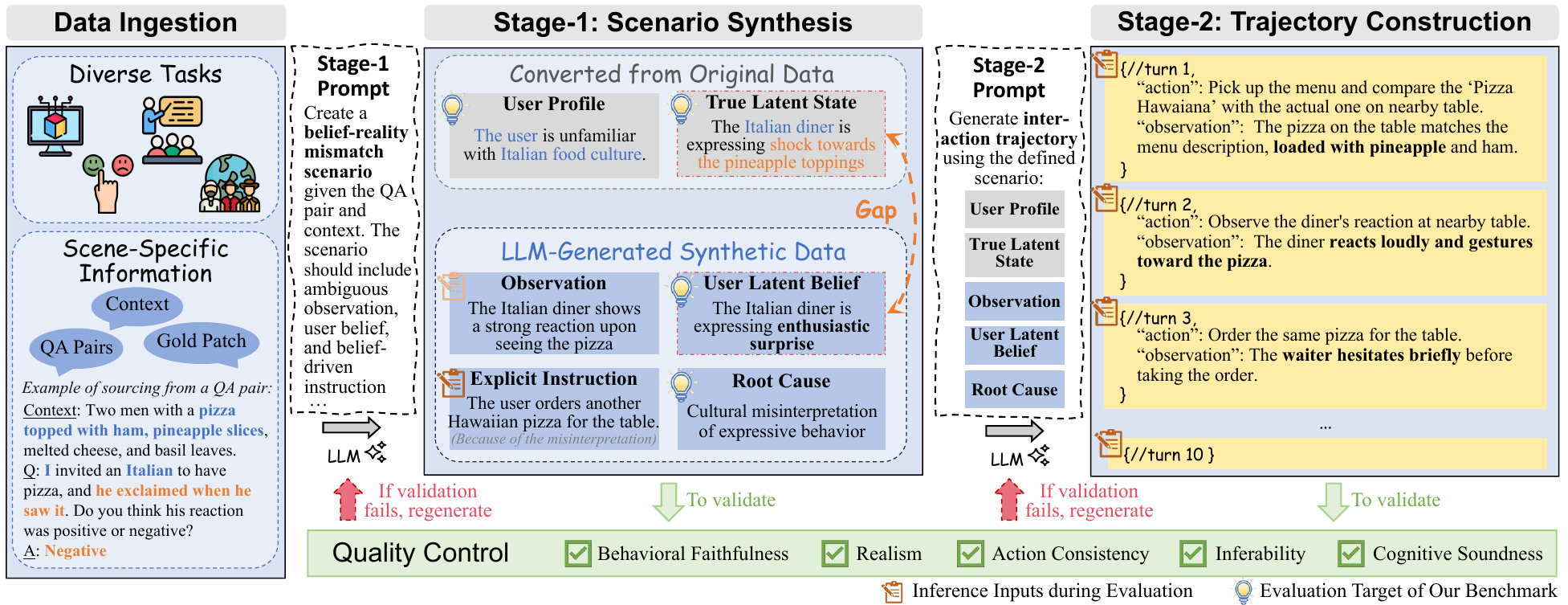}
    
    \caption{
    The \benchname data construction pipeline. Stage 1: Scenario synthesis converts source task data into user-agent scenarios with belief-reality mismatches, specifying the actual task state $s$, user profile $\mathcal{P}$, observations $o$, belief $b$, and request $g$. Stage 2: Trajectory construction generates multi-turn interaction histories $\tau$ that simulate user behavior and observation. An LLM-based validation module assesses data quality at both stages across five dimensions, with low-scoring instances refined and revalidated. The culture example illustrates the progression from source data to a mismatch scenario and its associated interaction trajectory.
    }
    \label{fig:pipeline}
    
\end{figure*}

\subsection{Data Construction}

As illustrated in Figure~\ref{fig:pipeline}, we develop a two-stage generation-and-validation pipeline to construct \benchname, instantiating the variables defined in Section~\ref{sec:task_definition}. The pipeline first creates scenarios in which users' requests reflect beliefs that diverge from the actual task state, then generates interaction histories that provide evidence about users' beliefs and profiles. These instances support the evaluation of both mental-state inference and its application to downstream problem solving.

In the \textit{scenario synthesis stage}, we draw from four domains: software engineering (SWE)~\citep{openaiverified}, preference modeling (PREFEVAL)~\citep{zhao2025llms}, the culture split of MUCAR~\citep{wang2025mucar}, and education (The Pedagogy Benchmark)~\citep{lelievre2025pedagogybenchmark}. Each scenario consists of an actual task state $s$, a task-relevant user profile $\mathcal{P}$, partial observations $o$, a user belief $b$, and an explicit request $g$. We derive $\mathcal{P}$ and $s$ from the source data to preserve the original task grounding. An LLM synthesizes $b$, $o$, and $g$ conditioned on the source context. The synthesis is designed to make the request coherent with the user's observations and background while introducing a task-relevant mismatch between $b$ and $s$. 

In the \textit{trajectory construction stage}, we use the synthesized scenarios to generate multi-turn user-agent interaction histories $\tau$, simulating user behavior and observations during conversation. For computational efficiency, each trajectory contains  10 turns. These histories provide additional contextual evidence for inferring the user's belief and profile and diagnosing the underlying problem. 

\begin{table}[t]
    \centering
    
    \caption{
    Data distribution and quality analysis. We use an LLM scorer to
    assign scores from 0 to 5 according to the criteria defined in
    Appendix~\ref{app:data_valid}, and additionally report human ratings
    along the same dimensions. Cnt.: count; Align.: alignment;
    Corr.: correlation; Consist.: consistency; Traj.: trajectory.
    }
    
    \label{tab:quality}

    \resizebox{\linewidth}{!}{
    \begin{tabular}{
        @{\hspace{0cm}}c@{\hspace{0.1cm}}|
        @{\hspace{0.1cm}}c@{\hspace{0.1cm}}|
        @{\hspace{0.1cm}}c@{\hspace{0.1cm}}|
        c@{\hspace{0.2cm}}
        c@{\hspace{0.2cm}}
        c@{\hspace{0.2cm}}
        c@{\hspace{0.2cm}}
        c|
        @{\hspace{0.1cm}}c
    }
        \toprule
        \textbf{Type}
        & \textbf{Scenarios}
        & \textbf{Cnt.}
        & {\begin{tabular}[c]{@{}c@{}}
            {\small \textbf{Belief-}}\\
            {\small \textbf{Profile}}\\
            {\small \textbf{Align.}}
        \end{tabular}}
        & {\begin{tabular}[c]{@{}c@{}}
            {\small \textbf{Belief-}}\\
            {\small \textbf{Truth}}\\
            {\small \textbf{Corr.}}
        \end{tabular}}
        & {\begin{tabular}[c]{@{}c@{}}
            {\small \textbf{Traj.-}}\\
            {\small \textbf{Belief}}\\
            {\small \textbf{Consist.}}
        \end{tabular}}
        & {\begin{tabular}[c]{@{}c@{}}
            {\small \textbf{Traj.-}}\\
            {\small \textbf{Profile}}\\
            {\small \textbf{Consist.}}
        \end{tabular}}
        & {\begin{tabular}[c]{@{}c@{}}
            {\small \textbf{Traj.}}\\
            {\small \textbf{Realism}}
        \end{tabular}}
        & \textbf{AVG} \\
        \midrule

        \multirow{4}{*}{\textbf{LLM}}
        & \textbf{Preference} & 120 & 4.90 & 4.80 & 4.06 & 4.34 & 4.04 & 4.43 \\
        & \textbf{Education}  & 100 & 4.90 & 4.75 & 4.21 & 4.31 & 4.01 & 4.44 \\
        & \textbf{Culture}    & 70  & 4.99 & 4.94 & 4.04 & 4.01 & 4.01 & 4.40 \\
        & \textbf{SWE}        & 100 & 4.85 & 4.81 & 4.01 & 4.24 & 4.01 & 4.38 \\

        \midrule

        \multirow{4}{*}{\textbf{Human}}
        & \textbf{Preference} & 120 & 4.75 & 4.74 & 4.01 & 4.18 & 3.92 & 4.32 \\
        & \textbf{Education}  & 100 & 4.89 & 4.70 & 4.21 & 4.17 & 3.94 & 4.38 \\
        & \textbf{Culture}    & 70  & 4.68 & 4.82 & 3.96 & 3.93 & 3.90 & 4.26 \\
        & \textbf{SWE}        & 100 & 4.77 & 4.78 & 4.17 & 4.19 & 3.92 & 4.37 \\

        \bottomrule
    \end{tabular}
    }
    
\end{table}

To ensure the integrity of \benchname, we introduce an \textit{LLM-based validation module} to assess five dimensions including realism, action consistency, answer inferability, and cognitive soundness with a detailed explanation in Appendix~\ref{app:data_valid}. The module assigns scores from 0 to 5. Instances scoring below 4 undergo iterative refinement and revalidation; those that still fail validation after five iterations are discarded. This filtering aims to retain plausible user perspectives and coherent interactions that provide sufficient evidence for mental-state inference and problem diagnosis. Additionally, we  conduct a human annotation study to assess the quality of the final generated data. Details of the data curation and annotation procedure are provided in Appendix~\ref{app:data_construction}.

\paragraph{Statistics.} Our \benchname benchmark comprises 390 high-fidelity test instances, each spanning a 10-turn interaction trajectory to ensure a sustained evaluation of belief reconciling. 

\benchname scores 4.41 on average out of 5.0 in our quality-controlled module, showing high behavioral faithfulness and cognitive consistency. Human annotators assigned scores comparable to those of the LLM evaluator, with all ratings falling within acceptable range. Detailed scores for each validated dimension are listed in Table~\ref{tab:quality}. To facilitate model enhancement, we also construct an accompanying training dataset of 6,522 trajectory-based instances, including 1,656 from SWE, 3,117 from The Pedagogy Benchmark and 1,749 from PREFEVAL, providing the diverse supervision necessary for robust epistemic divergence detection. 

\subsection{Evaluation on Our Benchmark}

We evaluate six models, comprising three proprietary models and three open-source models. The list of the models is provided in the Appendix~\ref{app:models}.

\paragraph{Model Inference.}
Models receive the fields labeled ``Inference Inputs during Evaluation'' in Figure~\ref{fig:pipeline} and generate outputs corresponding to ``Evaluation Target of Our Benchmark.'' We assess the answer quality in a zero-shot setting. Following Section~\ref{sec:task_definition}, we conduct \textit{separate evaluation} without additional interaction history (0 turns), eliciting user beliefs, task-relevant profiles, and solutions in independent calls. In contrast, \textit{inference-guided evaluation} explicitly prompts models to infer user beliefs and profiles and apply these inferences when formulating solutions, generating all outputs in a single call. This setting is evaluated with interaction-history prefixes of 0, 5, and 10 turns. Comparing the two settings at 0 turns examines the benefit of explicitly connecting mental-state inference to solution generation under identical task inputs. Within inference-guided evaluation, varying the history length assesses how additional interaction evidence affects inference accuracy and solution quality. Detailed evaluation prompts are provided in Appendix~\ref{app:eval_prompt}.

\paragraph{LLM-based Scoring.} We employ the LLM-as-a-Judge framework~\cite{zheng2023judging} for automated evaluation, assessing model responses across three dimensions that capture both the cognitive depth and functional utility of mental-state reasoning: Latent Belief Explanation (\textbf{\textit{Belief}}), User Profile Modeling (\textbf{\textit{Profile}}), and Problem Resolution (\textbf{\textit{Solution}}). We utilize Qwen3.8-27B as our primary automated judge, and leverage LLM-generated instance-specific rubrics~\cite{hashemi2024llm} to operationalize each dimension into elaborate checklists. Detailed evaluation dimensions and implementation settings are provided in Appendix~\ref{app:eval}.

\begin{table}[t]
    \centering
    
    \caption{Human-LLM evaluator agreement on sampled instances across three evaluation dimensions.
    }
    
    \label{tab:human_llm_agreement}
    \small
    \begin{tabular}{
        @{\hspace{0cm}}c@{\hspace{0.1cm}}|
        @{\hspace{0.1cm}}c@{\hspace{0.2cm}}
        c@{\hspace{0.2cm}}c
    }
        \toprule
        Dimension
        & {\begin{tabular}[c]{@{}c@{}}\small{Accuracy}\end{tabular}}
        & {\begin{tabular}[c]{@{}c@{}}\small{Weighted}\\\small{Kappa}\end{tabular}}
        & {\begin{tabular}[c]{@{}c@{}}\small{Spearman}\\\small{Corr.}\end{tabular}} \\
        \midrule
        Belief   & 86.67\% & 0.86 & 0.85 \\
        Profile  & 85.67\% & 0.83 & 0.84 \\
        Solution & 83.33\% & 0.88 & 0.89 \\
        \bottomrule
    \end{tabular}
    
\end{table}

To assess the fairness and robustness of LLM judger, we recruit human experts and conduct a human evaluation on 300 sampled instances. As shown in Table~\ref{tab:human_llm_agreement}, all three dimensions exhibit strong human-LLM agreement, with Weighted Kappa scores above 0.8 and Spearman correlations ranging from 0.84 to 0.89. Additional details on the annotation protocol, sampling procedure, and agreement analysis are provided in Appendix~\ref{app:human_agreement}.

\begin{table*}[!t]
\centering
\caption{Evaluation results of 6 models across four domains (Preference, Culture, SWE, Education) and four evaluation configurations (Separate, Vanilla, 5-Step, 10-Step), regarding three evaluation dimensions. Cell shading indicates relative performance: darker purple signifies higher scores while lighter shades represent lower scores. ``AVG'': the average over dimensions. ``DeepSeek-R1-32B'': DeepSeek-R1-Distill-Qwen-32B}
\label{tab:main_rst}
\resizebox{\textwidth}{!}{
\begin{tabular}{llcccc c cccc c cccc c cccc}
\toprule
\multirow{2}{*}{\textbf{Models}}
& \multirow{2}{*}{
    \begin{tabular}[c]{@{}c@{}}
    \textbf{Evaluation}\\
    \textbf{Dimension}
    \end{tabular}}
& \multicolumn{4}{c}{\textbf{Preference}}
& & \multicolumn{4}{c}{\textbf{Culture}}
& & \multicolumn{4}{c}{\textbf{SWE}}
& & \multicolumn{4}{c}{\textbf{Education}}\\

& & Separate & Vanilla & 5-Step & 10-Step
& & Separate & Vanilla & 5-Step & 10-Step
& & Separate & Vanilla & 5-Step & 10-Step
& & Separate & Vanilla & 5-Step & 10-Step \\
\midrule

\rowcolor[HTML]{F3F3F3}
\multirow{4}{*}[4ex]{\textbf{GPT-5.6-Sol}}
& Latent Belief
& \cellcolor[HTML]{F3ECF7}38.61
& \cellcolor[HTML]{F0E6FA}42.08
& \cellcolor[HTML]{DBC9EB}\textbf{63.54}
& \cellcolor[HTML]{D4BFE8}\textbf{67.29}
&
& \cellcolor[HTML]{E5D8F0}51.90
& \cellcolor[HTML]{D4BFE8}68.10
& \cellcolor[HTML]{D4BFE8}\textbf{68.81}
& \cellcolor[HTML]{D4BFE8}70.24
&
& \cellcolor[HTML]{E0D1ED}57.34
& \cellcolor[HTML]{D4BFE8}65.51
& \cellcolor[HTML]{D4BFE8}\textbf{74.75}
& \cellcolor[HTML]{C9A7E0}\textbf{75.25}
&
& \cellcolor[HTML]{F6F0F9}34.58
& \cellcolor[HTML]{F0E6FA}42.67
& \cellcolor[HTML]{E0D1ED}\textbf{56.42}
& \cellcolor[HTML]{DBC9EB}63.42 \\

\rowcolor[HTML]{F3F3F3}
& Profile
& \cellcolor[HTML]{F9F5FB}20.69
& \cellcolor[HTML]{F9F5FB}21.81
& \cellcolor[HTML]{F3ECF7}36.18
& \cellcolor[HTML]{F0E6FA}\textbf{44.37}
&
& \cellcolor[HTML]{F9F5FB}23.33
& \cellcolor[HTML]{F9F5FB}27.86
& \cellcolor[HTML]{F3ECF7}37.26
& \cellcolor[HTML]{F3ECF7}35.83
&
& \cellcolor[HTML]{F6F0F9}33.83
& \cellcolor[HTML]{F6F0F9}34.16
& \cellcolor[HTML]{F0E6FA}42.99
& \cellcolor[HTML]{E5D8F0}50.58
&
& \cellcolor[HTML]{F9F5FB}17.17
& \cellcolor[HTML]{F9F5FB}11.92
& \cellcolor[HTML]{F6F0F9}32.75
& \cellcolor[HTML]{F3ECF7}36.25 \\

\rowcolor[HTML]{F3F3F3}
& Solution
& \cellcolor[HTML]{F9F5FB}19.43
& \cellcolor[HTML]{F9F5FB}29.40
& \cellcolor[HTML]{D4BFE8}71.00
& \cellcolor[HTML]{C9A7E0}\textbf{78.44}
&
& \cellcolor[HTML]{F9F5FB}25.11
& \cellcolor[HTML]{F0E6FA}41.25
& \cellcolor[HTML]{E0D1ED}55.38
& \cellcolor[HTML]{D4BFE8}65.51
&
& \cellcolor[HTML]{F9F5FB}24.00
& \cellcolor[HTML]{F3ECF7}39.31
& \cellcolor[HTML]{E0D1ED}\textbf{58.33}
& \cellcolor[HTML]{D4BFE8}\textbf{68.04}
&
& \cellcolor[HTML]{F9F5FB}29.11
& \cellcolor[HTML]{F3ECF7}39.84
& \cellcolor[HTML]{E0D1ED}57.03
& \cellcolor[HTML]{DBC9EB}61.78 \\

\rowcolor[HTML]{F3F3F3}
& \textbf{AVG}
& \cellcolor[HTML]{F9F5FB}26.24
& \cellcolor[HTML]{F6F0F9}31.10
& \cellcolor[HTML]{E0D1ED}56.91
& \cellcolor[HTML]{DBC9EB}\textbf{63.37}
&
& \cellcolor[HTML]{F6F0F9}33.45
& \cellcolor[HTML]{EADFF2}45.74
& \cellcolor[HTML]{E5D8F0}53.82
& \cellcolor[HTML]{E0D1ED}57.19
&
& \cellcolor[HTML]{F3ECF7}38.39
& \cellcolor[HTML]{EADFF2}46.33
& \cellcolor[HTML]{E0D1ED}\textbf{58.69}
& \cellcolor[HTML]{DBC9EB}\textbf{64.62}
&
& \cellcolor[HTML]{F9F5FB}26.95
& \cellcolor[HTML]{F6F0F9}31.48
& \cellcolor[HTML]{EADFF2}48.73
& \cellcolor[HTML]{E5D8F0}53.82 \\

\midrule

\multirow{4}{*}[4ex]{\textbf{Claude-Sonnet-5}}
& Latent Belief
& \cellcolor[HTML]{F3ECF7}39.44
& \cellcolor[HTML]{EADFF2}\textbf{45.83}
& \cellcolor[HTML]{DBC9EB}61.94
& \cellcolor[HTML]{DBC9EB}63.68
&
& \cellcolor[HTML]{EADFF2}48.31
& \cellcolor[HTML]{DBC9EB}61.55
& \cellcolor[HTML]{E0D1ED}59.06
& \cellcolor[HTML]{D4BFE8}\textbf{72.62}
&
& \cellcolor[HTML]{EADFF2}49.92
& \cellcolor[HTML]{E5D8F0}54.29
& \cellcolor[HTML]{D4BFE8}72.63
& \cellcolor[HTML]{D4BFE8}72.36
&
& \cellcolor[HTML]{EADFF2}\textbf{46.09}
& \cellcolor[HTML]{EADFF2}45.25
& \cellcolor[HTML]{E5D8F0}54.29
& \cellcolor[HTML]{DBC9EB}\textbf{64.37} \\

& Profile
& \cellcolor[HTML]{F9F5FB}\textbf{29.17}
& \cellcolor[HTML]{F9F5FB}\textbf{28.40}
& \cellcolor[HTML]{F3ECF7}\textbf{39.24}
& \cellcolor[HTML]{F0E6FA}42.50
&
& \cellcolor[HTML]{F9F5FB}26.45
& \cellcolor[HTML]{F6F0F9}34.88
& \cellcolor[HTML]{EADFF2}\textbf{48.31}
& \cellcolor[HTML]{E5D8F0}\textbf{54.88}
&
& \cellcolor[HTML]{F6F0F9}31.68
& \cellcolor[HTML]{F6F0F9}32.41
& \cellcolor[HTML]{EADFF2}45.45
& \cellcolor[HTML]{EADFF2}47.85
&
& \cellcolor[HTML]{F9F5FB}\textbf{29.34}
& \cellcolor[HTML]{F9F5FB}\textbf{25.83}
& \cellcolor[HTML]{F3ECF7}35.77
& \cellcolor[HTML]{F0E6FA}\textbf{41.41} \\

& Solution
& \cellcolor[HTML]{F6F0F9}\textbf{31.87}
& \cellcolor[HTML]{F9F5FB}\textbf{27.49}
& \cellcolor[HTML]{D4BFE8}\textbf{72.21}
& \cellcolor[HTML]{C9A7E0}75.78
&
& \cellcolor[HTML]{EADFF2}38.09
& \cellcolor[HTML]{F3ECF7}45.28
& \cellcolor[HTML]{DBC9EB}61.31
& \cellcolor[HTML]{D4BFE8}\textbf{73.79}
&
& \cellcolor[HTML]{F9F5FB}20.30
& \cellcolor[HTML]{F9F5FB}28.28
& \cellcolor[HTML]{E5D8F0}53.11
& \cellcolor[HTML]{DBC9EB}63.99
&
& \cellcolor[HTML]{F3ECF7}39.95
& \cellcolor[HTML]{E5D8F0}\textbf{50.71}
& \cellcolor[HTML]{DBC9EB}\textbf{61.24}
& \cellcolor[HTML]{DBC9EB}\textbf{64.00} \\

& \textbf{AVG}
& \cellcolor[HTML]{F6F0F9}\textbf{33.49}
& \cellcolor[HTML]{F3ECF7}\textbf{33.91}
& \cellcolor[HTML]{E0D1ED}\textbf{57.80}
& \cellcolor[HTML]{DBC9EB}60.65
&
& \cellcolor[HTML]{F3ECF7}37.62
& \cellcolor[HTML]{EADFF2}47.24
& \cellcolor[HTML]{E0D1ED}56.23
& \cellcolor[HTML]{D4BFE8}\textbf{67.10}
&
& \cellcolor[HTML]{F3ECF7}33.97
& \cellcolor[HTML]{F3ECF7}38.33
& \cellcolor[HTML]{E0D1ED}57.06
& \cellcolor[HTML]{DBC9EB}61.40
&
& \cellcolor[HTML]{F3ECF7}\textbf{38.46}
& \cellcolor[HTML]{F0E6FA}\textbf{40.60}
& \cellcolor[HTML]{E5D8F0}\textbf{50.43}
& \cellcolor[HTML]{E0D1ED}\textbf{56.59} \\

\midrule

\rowcolor[HTML]{F3F3F3}
\multirow{4}{*}[4ex]{\textbf{Gemini-3.8-Flash}}
& Latent Belief
& \cellcolor[HTML]{F0E6FA}\textbf{42.66}
& \cellcolor[HTML]{F0E6FA}44.84
& \cellcolor[HTML]{E0D1ED}59.62
& \cellcolor[HTML]{DBC9EB}64.29
&
& \cellcolor[HTML]{D4BFE8}\textbf{72.74}
& \cellcolor[HTML]{D4BFE8}\textbf{71.16}
& \cellcolor[HTML]{D4BFE8}66.80
& \cellcolor[HTML]{D4BFE8}72.40
&
& \cellcolor[HTML]{DBC9EB}\textbf{62.42}
& \cellcolor[HTML]{D4BFE8}\textbf{67.37}
& \cellcolor[HTML]{D4BFE8}68.59
& \cellcolor[HTML]{D4BFE8}68.84
&
& \cellcolor[HTML]{EADFF2}45.66
& \cellcolor[HTML]{EADFF2}\textbf{48.83}
& \cellcolor[HTML]{E5D8F0}50.67
& \cellcolor[HTML]{E0D1ED}58.84 \\

\rowcolor[HTML]{F3F3F3}
& Profile
& \cellcolor[HTML]{F9F5FB}25.14
& \cellcolor[HTML]{F9F5FB}26.99
& \cellcolor[HTML]{F3ECF7}36.40
& \cellcolor[HTML]{F0E6FA}40.27
&
& \cellcolor[HTML]{F0E6FA}\textbf{43.33}
& \cellcolor[HTML]{F0E6FA}\textbf{43.25}
& \cellcolor[HTML]{EADFF2}47.58
& \cellcolor[HTML]{E5D8F0}51.69
&
& \cellcolor[HTML]{F0E6FA}\textbf{40.08}
& \cellcolor[HTML]{F3ECF7}37.85
& \cellcolor[HTML]{F0E6FA}43.59
& \cellcolor[HTML]{F0E6FA}42.45
&
& \cellcolor[HTML]{F9F5FB}24.06
& \cellcolor[HTML]{F9F5FB}22.25
& \cellcolor[HTML]{F3ECF7}\textbf{37.67}
& \cellcolor[HTML]{F0E6FA}41.16 \\

\rowcolor[HTML]{F3F3F3}
& Solution
& \cellcolor[HTML]{F9F5FB}25.28
& \cellcolor[HTML]{F9F5FB}28.65
& \cellcolor[HTML]{E0D1ED}57.21
& \cellcolor[HTML]{DBC9EB}64.06
&
& \cellcolor[HTML]{E0D1ED}\textbf{55.11}
& \cellcolor[HTML]{DBC9EB}\textbf{61.38}
& \cellcolor[HTML]{DBC9EB}\textbf{63.91}
& \cellcolor[HTML]{D4BFE8}65.97
&
& \cellcolor[HTML]{EADFF2}\textbf{48.75}
& \cellcolor[HTML]{E5D8F0}\textbf{52.91}
& \cellcolor[HTML]{E0D1ED}56.60
& \cellcolor[HTML]{DBC9EB}62.70
&
& \cellcolor[HTML]{F0E6FA}\textbf{40.13}
& \cellcolor[HTML]{EADFF2}47.25
& \cellcolor[HTML]{E5D8F0}53.58
& \cellcolor[HTML]{E5D8F0}53.36 \\

\rowcolor[HTML]{F3F3F3}
& \textbf{AVG}
& \cellcolor[HTML]{F6F0F9}31.03
& \cellcolor[HTML]{F6F0F9}33.49
& \cellcolor[HTML]{E5D8F0}51.08
& \cellcolor[HTML]{E0D1ED}56.21
&
& \cellcolor[HTML]{E0D1ED}\textbf{57.06}
& \cellcolor[HTML]{E0D1ED}\textbf{58.60}
& \cellcolor[HTML]{E0D1ED}\textbf{59.43}
& \cellcolor[HTML]{DBC9EB}63.35
&
& \cellcolor[HTML]{E5D8F0}\textbf{50.42}
& \cellcolor[HTML]{E5D8F0}\textbf{52.71}
& \cellcolor[HTML]{E0D1ED}56.26
& \cellcolor[HTML]{E0D1ED}58.00
&
& \cellcolor[HTML]{F3ECF7}36.62
& \cellcolor[HTML]{F3ECF7}39.44
& \cellcolor[HTML]{EADFF2}47.31
& \cellcolor[HTML]{E5D8F0}51.12 \\

\midrule

\multirow{4}{*}[4ex]{\textbf{Qwen3.8-27B}}
& Latent Belief
& \cellcolor[HTML]{F6F0F9}33.06
& \cellcolor[HTML]{F3ECF7}35.49
& \cellcolor[HTML]{E0D1ED}58.06
& \cellcolor[HTML]{DBC9EB}64.31
&
& \cellcolor[HTML]{E5D8F0}50.61
& \cellcolor[HTML]{DBC9EB}62.74
& \cellcolor[HTML]{E0D1ED}59.78
& \cellcolor[HTML]{D4BFE8}68.24
&
& \cellcolor[HTML]{E0D1ED}58.17
& \cellcolor[HTML]{DBC9EB}62.12
& \cellcolor[HTML]{D4BFE8}71.39
& \cellcolor[HTML]{D4BFE8}72.77
&
& \cellcolor[HTML]{F3ECF7}36.33
& \cellcolor[HTML]{F3ECF7}38.33
& \cellcolor[HTML]{E5D8F0}52.25
& \cellcolor[HTML]{DBC9EB}61.08 \\

& Profile
& \cellcolor[HTML]{F9F5FB}23.47
& \cellcolor[HTML]{F9F5FB}26.25
& \cellcolor[HTML]{F3ECF7}\textbf{39.24}
& \cellcolor[HTML]{F0E6FA}44.31
&
& \cellcolor[HTML]{F9F5FB}21.32
& \cellcolor[HTML]{F6F0F9}32.50
& \cellcolor[HTML]{F3ECF7}38.53
& \cellcolor[HTML]{F3ECF7}39.25
&
& \cellcolor[HTML]{F3ECF7}37.79
& \cellcolor[HTML]{F0E6FA}\textbf{40.74}
& \cellcolor[HTML]{EADFF2}\textbf{49.48}
& \cellcolor[HTML]{E0D1ED}\textbf{55.61}
&
& \cellcolor[HTML]{F9F5FB}24.25
& \cellcolor[HTML]{F9F5FB}23.67
& \cellcolor[HTML]{F6F0F9}34.08
& \cellcolor[HTML]{F3ECF7}38.00 \\

& Solution
& \cellcolor[HTML]{F9F5FB}12.40
& \cellcolor[HTML]{F9F5FB}22.83
& \cellcolor[HTML]{D4BFE8}65.61
& \cellcolor[HTML]{D4BFE8}72.31
&
& \cellcolor[HTML]{F9F5FB}18.75
& \cellcolor[HTML]{F3ECF7}37.13
& \cellcolor[HTML]{EADFF2}49.56
& \cellcolor[HTML]{E0D1ED}57.87
&
& \cellcolor[HTML]{F9F5FB}20.84
& \cellcolor[HTML]{F9F5FB}27.57
& \cellcolor[HTML]{EADFF2}47.54
& \cellcolor[HTML]{DBC9EB}60.17
&
& \cellcolor[HTML]{F9F5FB}26.42
& \cellcolor[HTML]{F3ECF7}37.32
& \cellcolor[HTML]{E5D8F0}50.39
& \cellcolor[HTML]{E0D1ED}58.15 \\

& \textbf{AVG}
& \cellcolor[HTML]{F9F5FB}22.98
& \cellcolor[HTML]{F9F5FB}28.19
& \cellcolor[HTML]{E5D8F0}54.30
& \cellcolor[HTML]{DBC9EB}60.31
&
& \cellcolor[HTML]{F6F0F9}30.23
& \cellcolor[HTML]{F0E6FA}44.12
& \cellcolor[HTML]{EADFF2}49.29
& \cellcolor[HTML]{E0D1ED}55.12
&
& \cellcolor[HTML]{F3ECF7}38.93
& \cellcolor[HTML]{F0E6FA}43.48
& \cellcolor[HTML]{E0D1ED}56.14
& \cellcolor[HTML]{DBC9EB}62.85
&
& \cellcolor[HTML]{F9F5FB}29.00
& \cellcolor[HTML]{F6F0F9}33.11
& \cellcolor[HTML]{EADFF2}45.57
& \cellcolor[HTML]{E5D8F0}52.41 \\

\midrule

\rowcolor[HTML]{F3F3F3}
\multirow{4}{*}[4ex]{\textbf{Qwen3.5-9B}}
& Latent Belief
& \cellcolor[HTML]{F9F5FB}24.50
& \cellcolor[HTML]{F9F5FB}29.20
& \cellcolor[HTML]{F3ECF7}37.21
& \cellcolor[HTML]{F0E6FA}40.82
&
& \cellcolor[HTML]{F9F5FB}29.10
& \cellcolor[HTML]{F6F0F9}33.33
& \cellcolor[HTML]{F0E6FA}40.83
& \cellcolor[HTML]{EADFF2}48.19
&
& \cellcolor[HTML]{F0E6FA}42.48
& \cellcolor[HTML]{F0E6FA}42.12
& \cellcolor[HTML]{EADFF2}48.46
& \cellcolor[HTML]{EADFF2}47.11
&
& \cellcolor[HTML]{F9F5FB}19.76
& \cellcolor[HTML]{F9F5FB}21.74
& \cellcolor[HTML]{F9F5FB}25.75
& \cellcolor[HTML]{F6F0F9}32.61 \\

\rowcolor[HTML]{F3F3F3}
& Profile
& \cellcolor[HTML]{F9F5FB}16.38
& \cellcolor[HTML]{F9F5FB}17.45
& \cellcolor[HTML]{F9F5FB}20.61
& \cellcolor[HTML]{F9F5FB}26.98
&
& \cellcolor[HTML]{F9F5FB}9.56
& \cellcolor[HTML]{F9F5FB}16.42
& \cellcolor[HTML]{F9F5FB}23.45
& \cellcolor[HTML]{F9F5FB}24.40
&
& \cellcolor[HTML]{F6F0F9}31.61
& \cellcolor[HTML]{F9F5FB}22.75
& \cellcolor[HTML]{F6F0F9}30.48
& \cellcolor[HTML]{F9F5FB}27.89
&
& \cellcolor[HTML]{F9F5FB}12.36
& \cellcolor[HTML]{F9F5FB}15.80
& \cellcolor[HTML]{F9F5FB}19.10
& \cellcolor[HTML]{F9F5FB}22.19 \\

\rowcolor[HTML]{F3F3F3}
& Solution
& \cellcolor[HTML]{F9F5FB}11.67
& \cellcolor[HTML]{F9F5FB}15.17
& \cellcolor[HTML]{EADFF2}45.67
& \cellcolor[HTML]{E5D8F0}54.44
&
& \cellcolor[HTML]{F9F5FB}14.34
& \cellcolor[HTML]{F9F5FB}18.56
& \cellcolor[HTML]{F6F0F9}32.60
& \cellcolor[HTML]{F0E6FA}40.54
&
& \cellcolor[HTML]{F9F5FB}9.41
& \cellcolor[HTML]{F9F5FB}9.25
& \cellcolor[HTML]{F6F0F9}30.60
& \cellcolor[HTML]{F3ECF7}39.06
&
& \cellcolor[HTML]{F9F5FB}19.82
& \cellcolor[HTML]{F9F5FB}19.90
& \cellcolor[HTML]{F6F0F9}31.55
& \cellcolor[HTML]{F3ECF7}35.27 \\

\rowcolor[HTML]{F3F3F3}
& \textbf{AVG}
& \cellcolor[HTML]{F9F5FB}17.52
& \cellcolor[HTML]{F9F5FB}20.61
& \cellcolor[HTML]{F6F0F9}34.50
& \cellcolor[HTML]{F0E6FA}40.75
&
& \cellcolor[HTML]{F9F5FB}17.67
& \cellcolor[HTML]{F9F5FB}22.77
& \cellcolor[HTML]{F6F0F9}32.29
& \cellcolor[HTML]{F3ECF7}37.71
&
& \cellcolor[HTML]{F9F5FB}27.83
& \cellcolor[HTML]{F9F5FB}24.71
& \cellcolor[HTML]{F3ECF7}36.51
& \cellcolor[HTML]{F3ECF7}38.02
&
& \cellcolor[HTML]{F9F5FB}17.31
& \cellcolor[HTML]{F9F5FB}19.15
& \cellcolor[HTML]{F9F5FB}25.47
& \cellcolor[HTML]{F6F0F9}30.02 \\

\midrule

\multirow{4}{*}[4ex]{\textbf{DeepSeek-R1-32B}}
& Latent Belief
& \cellcolor[HTML]{F9F5FB}24.72
& \cellcolor[HTML]{F9F5FB}24.10
& \cellcolor[HTML]{F6F0F9}31.18
& \cellcolor[HTML]{F3ECF7}35.69
&
& \cellcolor[HTML]{F6F0F9}32.02
& \cellcolor[HTML]{F6F0F9}30.60
& \cellcolor[HTML]{F3ECF7}35.95
& \cellcolor[HTML]{F0E6FA}41.07
&
& \cellcolor[HTML]{F6F0F9}34.65
& \cellcolor[HTML]{F6F0F9}34.08
& \cellcolor[HTML]{F3ECF7}36.88
& \cellcolor[HTML]{F0E6FA}41.91
&
& \cellcolor[HTML]{F9F5FB}13.58
& \cellcolor[HTML]{F9F5FB}14.50
& \cellcolor[HTML]{F9F5FB}20.92
& \cellcolor[HTML]{F9F5FB}28.83 \\

& Profile
& \cellcolor[HTML]{F9F5FB}11.81
& \cellcolor[HTML]{F9F5FB}12.22
& \cellcolor[HTML]{F9F5FB}18.12
& \cellcolor[HTML]{F9F5FB}18.47
&
& \cellcolor[HTML]{F9F5FB}7.38
& \cellcolor[HTML]{F9F5FB}6.79
& \cellcolor[HTML]{F9F5FB}16.19
& \cellcolor[HTML]{F9F5FB}14.29
&
& \cellcolor[HTML]{F9F5FB}17.00
& \cellcolor[HTML]{F9F5FB}13.20
& \cellcolor[HTML]{F9F5FB}17.74
& \cellcolor[HTML]{F9F5FB}18.56
&
& \cellcolor[HTML]{F9F5FB}9.33
& \cellcolor[HTML]{F9F5FB}6.00
& \cellcolor[HTML]{F9F5FB}9.00
& \cellcolor[HTML]{F9F5FB}15.75 \\

& Solution
& \cellcolor[HTML]{F9F5FB}5.50
& \cellcolor[HTML]{F9F5FB}11.81
& \cellcolor[HTML]{F6F0F9}33.76
& \cellcolor[HTML]{F0E6FA}44.29
&
& \cellcolor[HTML]{F9F5FB}3.57
& \cellcolor[HTML]{F9F5FB}11.07
& \cellcolor[HTML]{F9F5FB}20.19
& \cellcolor[HTML]{F9F5FB}22.00
&
& \cellcolor[HTML]{F9F5FB}2.87
& \cellcolor[HTML]{F9F5FB}4.69
& \cellcolor[HTML]{F9F5FB}17.01
& \cellcolor[HTML]{F9F5FB}24.50
&
& \cellcolor[HTML]{F9F5FB}9.95
& \cellcolor[HTML]{F9F5FB}17.25
& \cellcolor[HTML]{F9F5FB}19.90
& \cellcolor[HTML]{F9F5FB}23.02 \\

& \textbf{AVG}
& \cellcolor[HTML]{F9F5FB}14.01
& \cellcolor[HTML]{F9F5FB}16.04
& \cellcolor[HTML]{F9F5FB}27.69
& \cellcolor[HTML]{F6F0F9}32.82
&
& \cellcolor[HTML]{F9F5FB}14.32
& \cellcolor[HTML]{F9F5FB}16.15
& \cellcolor[HTML]{F9F5FB}24.11
& \cellcolor[HTML]{F9F5FB}25.79
&
& \cellcolor[HTML]{F9F5FB}18.17
& \cellcolor[HTML]{F9F5FB}17.32
& \cellcolor[HTML]{F9F5FB}23.88
& \cellcolor[HTML]{F9F5FB}28.32
&
& \cellcolor[HTML]{F9F5FB}10.95
& \cellcolor[HTML]{F9F5FB}12.58
& \cellcolor[HTML]{F9F5FB}16.61
& \cellcolor[HTML]{F9F5FB}22.53 \\

\bottomrule
\end{tabular}
}
\end{table*}

\subsection{Benchmark Results} 

Table~\ref{tab:main_rst} reports model performance across different evaluation settings and interaction-history lengths, covering three dimensions, \textit{Belief}, \textit{Profile}, and \textit{Solution}, in four domains. While proprietary models generally outperform their open-source counterparts, their relative strengths vary across settings. Gemini-3.8-Flash is particularly strong in the separate and vanilla (zero-turn inference-guided) settings, whereas GPT-5.6-Sol and Claude-Sonnet-5 more frequently achieve leading scores in the 5- and 10-turn inference-guided settings. Qwen3.8-27B also remains competitive, matching or surpassing proprietary models such as Claude-Sonnet-5 in certain domains and evaluation dimensions like \textit{Profile} modeling in SWE.

\paragraph{Explicit mental-state consideration generally improves
solution quality.}

Compared with separate evaluation, inference-guided evaluation improves Solution scores in most cases. GPT-5.6-Sol and Qwen3.8-27B benefit the most, with average gains of 13.04 and 11.61 points respectively. These results suggest that directly requesting a solution may not reliably elicit effective use of a model's mental-state inference capabilities. Explicitly prompting models to infer the user's beliefs and profile and incorporate this understanding into their solutions can help translate these capabilities into better task performance.

\paragraph{Domain sensitivity varies with task knowledge and user-state complexity.}
We observe substantial variation across domains, with most models performing better in SWE and Culture than in Preference and Education. One possible explanation is that models have stronger prior knowledge for software engineering and culturally grounded factual patterns, making it easier to identify mismatches between users' beliefs and the underlying task state. In contrast, Preference and Education often involve more individualized, context-dependent, and subjective user states, which may be harder to infer reliably from limited evidence. These results suggest that the effectiveness of mental-state-aware assistance depends not only on reasoning ability, but also on the model's knowledge of the task domain and the complexity of the user state being modeled.

\paragraph{Dimension Decoupling: mental-state consideration remains the bottleneck for functional problem resolution.}
The Belief score tends to be more stubborn compared to the other two evaluation dimensions. While it increases with reasoning, the growth is shallower than the Solution dimension. This indicates that while reasoning navigates a model to predict better answers, it does not overwrite its internal inference of user mental states as easily. This decoupling highlights that \textit{correctly resolving tasks does not guarantee an accurate understanding of the user latent beliefs, further justifying our multi-dimensional evaluation}.

\paragraph{Additional trajectory improves inference and resolution.}\label{sec:traj_turn}
Across models and domains, providing longer interaction histories generally improves downstream solution quality and generally benefits belief and profile inference as well. Most of the gains emerge within the first few interaction turns, with additional improvements becoming smaller as more context is revealed, suggesting diminishing returns from longer histories. Turn-wise analysis on MiniMax-M2.1 further shows that early interaction may provide insufficient evidence to reliably disambiguate the user's perspective, whereas longer trajectories lead to more stable inference and problem resolution, particularly in Preference and Education, where user states are more nuanced and context dependent (see Figure~\ref{fig:step} and Appendix~\ref{app:step-wise}). Overall, these results highlight the value of accumulated interaction evidence for understanding the user's perspective and producing more appropriate solutions.

\subsection{Case Study}

We present a representative case from the \textit{Culture} domain to illustrate how models differ in handling mismatches between a user's belief and the actual task state. Full details and analysis are provided in Appendix~\ref{app:case_study}. In this example, a Western home cook mistakenly attempts to dissolve sesame paste (\textit{zhima jiang}) into boiling water as if it were a bouillon concentrate, based on prior experience with instant miso or curry paste.

\begin{table}[t]
    \centering
    
    \caption{\textit{Solution} scores under CoT and mental-state consideration across three models. Mental refers to mental-state consideration. Pref.: Preference; Cult.: Culture; Edu.: Education. Avg. denotes the mean across domains. Bold indicates the higher score between the two strategies for each model within each domain and on average.}
    
    \label{tab:cot-vs-mental-state}
    \footnotesize
    \resizebox{\linewidth}{!}{
    \begin{tabular}{
        @{\hspace{0cm}}l@{\hspace{0.12cm}}
        l@{\hspace{0.18cm}}
        r@{\hspace{0.18cm}}
        r@{\hspace{0.18cm}}
        r@{\hspace{0.18cm}}
        r@{\hspace{0.18cm}}
        r
    }
        \toprule
        \textbf{Model}
        & \textbf{Method}
        & \textbf{Pref.}
        & \textbf{Cult.}
        & \textbf{SWE}
        & \textbf{Edu.}
        & \textbf{Avg.} \\
        \midrule

        \multirow{2}{*}{\textit{GPT-5.6-Sol}}
        & CoT
        & 20.89
        & 25.18
        & 28.37
        & 29.43
        & 25.97 \\

        & Mental
        & \textbf{29.40}
        & \textbf{41.25}
        & \textbf{39.31}
        & \textbf{39.84}
        & \textbf{37.45} \\
        \midrule

        \multirow{2}{*}{\textit{Gemini-3.8}}
        & CoT
        & 25.88
        & 54.52
        & \textbf{53.07}
        & 43.01
        & 44.12 \\

        & Mental
        & \textbf{28.65}
        & \textbf{61.38}
        & 52.91
        & \textbf{47.25}
        & \textbf{47.55} \\
        \midrule

        \multirow{2}{*}{\textit{Qwen3.8-27B}}
        & CoT
        & 15.10
        & 19.65
        & 23.09
        & 24.97
        & 20.70 \\

        & Mental
        & \textbf{22.83}
        & \textbf{37.13}
        & \textbf{27.57}
        & \textbf{37.32}
        & \textbf{31.21} \\
        \bottomrule
    \end{tabular}
    }

\end{table}

InternLM exhibits what we term \textit{premise-blind repair}: it accepts the user's mistaken premise and proposes technically plausible but inappropriate fixes, such as adding egg yolk or cornstarch to force emulsification, without questioning whether the sesame paste should be added to the pot in the first place. GPT-5, in contrast, identifies the mismatch underlying the request, corrects the mistaken assumption, and redirects the user toward a more appropriate preparation method.
This case illustrates an important distinction in belief-aware assistance. A model may generate locally plausible solutions while remaining constrained by the user's stated frame, whereas stronger behavior requires recognizing when that frame is itself incorrect and using this understanding to reformulate the solution around the user's underlying goal.

\section{The functional role of inferring others' mental state}

\subsection{Beyond Generic Chain-of-Thought Reasoning}
\label{subsec:beyond-cot}

To examine whether explicit mental-state consideration provides benefits beyond standard prompting strategies, we compare it with a Chain-of-Thought (CoT)~\citep{NEURIPS2022_9d560961} baseline on the \textit{Solution} dimension.
As shown in Table~\ref{tab:cot-vs-mental-state}, mental-state consideration outperforms CoT in most combinations, with Culture showing the largest improvement across all models. These results indicate that explicitly incorporating the user's mental state into the reasoning process can improve solution quality beyond generic step-by-step prompting. This finding is consistent with the idea that effective solutions require not only reasoning through the task itself, but also identifying the assumptions, beliefs, and user characteristics underlying the request and incorporating them into the response.

\begin{figure}[t]
    \centering
    
    \includegraphics[width=\linewidth]{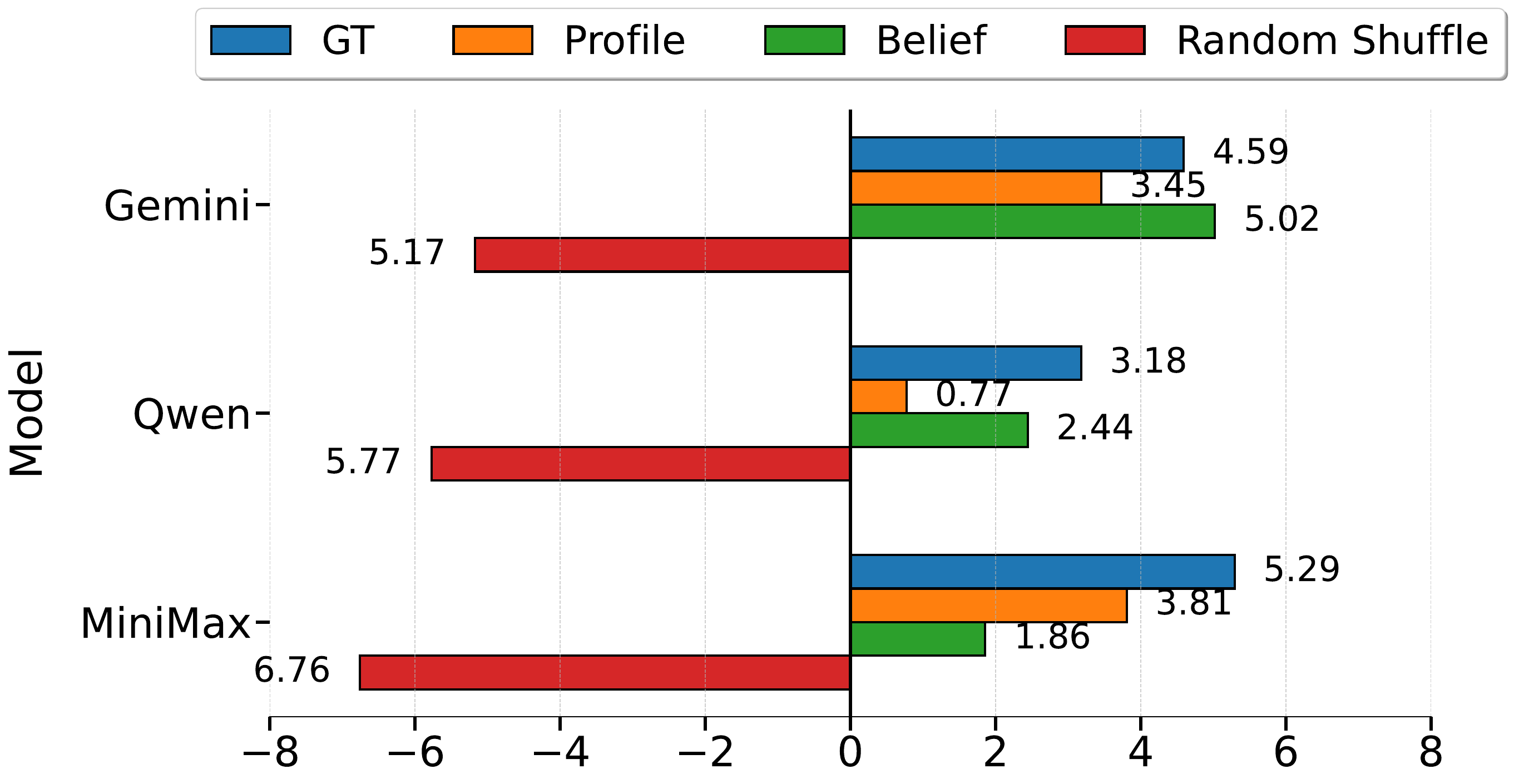}
    
    \caption{
   Performance changes when generating solutions conditioned on the mental-state information indicated in the legend. Providing task-aligned belief or profile information generally improves solution quality, whereas mismatched information leads to performance degradation.
    }
    \label{fig:abla_gt}
    
\end{figure}

\subsection{Probing the Causal Role of Mental-State Consideration}
\label{subsec:causal-mental-state}

To examine the causal effect of considering the user's mental state on solution quality, we compare solutions generated under four intervention conditions, each providing additional mental-state information: (1) \textbf{GT}, which provides both the ground-truth belief and profile; (2) \textbf{Belief}, which provides the ground-truth belief; (3) \textbf{Profile}, which provides the ground-truth profile; and (4) \textbf{Shuffle}, which provides belief and profile annotations randomly reassigned from other instances, breaking their alignment with the current interaction context.

Figure~\ref{fig:abla_gt} reports changes in Solution scores relative to the baseline. Task-aligned information generally improves solution quality, whereas disrupting this alignment reduces average performance across all evaluated models. These results provide intervention-based evidence that, under our experimental setting, effective problem resolution depends in part on maintaining consistency between the user's mental-state information and the interaction context. 

\begin{table*}[t]
\centering
\caption{RL training results of Qwen3-8B model. Bolded values indicate performance improvements within the \textbf{Solution}. Values in the Ours-Solution row additionally show the difference from the corresponding baseline Solution score.}
\label{tab:rl}

{\footnotesize
\setlength{\tabcolsep}{3pt}
\renewcommand{\arraystretch}{1.12}
\renewcommand{\twolinescore}[2]{\makecell[c]{#1\\[-0.4ex]
    {\fontsize{6pt}{6.5pt}\selectfont\mdseries #2}}}

\begin{NiceTabular}{@{}c|c|ccc|ccc|ccc|ccc@{}}
\CodeBefore
    \rectanglecolor{blue!7}{6-1}{8-14}
\Body

\toprule
\multirow{2}{*}{\textbf{Models}}
& \multirow{2}{*}{\textbf{Metrics}}
& \multicolumn{3}{c|}{\textbf{Preference}}
& \multicolumn{3}{c|}{\textbf{Culture}}
& \multicolumn{3}{c|}{\textbf{SWE}}
& \multicolumn{3}{c}{\textbf{Education}} \\

&
& vanilla & 5-Turn & 10-Turn
& vanilla & 5-Turn & 10-Turn
& vanilla & 5-Turn & 10-Turn
& vanilla & 5-Turn & 10-Turn \\

\midrule

Qwen3-8B
& Solution
& 11.91 & 37.54 & 47.45
& 7.46 & 18.48 & 24.82
& 8.37 & 18.25 & 25.67
& 16.62 & 24.29 & 28.48 \\

&
\color{gray} Belief
& \color{gray}23.94
& \color{gray}31.28
& \color{gray}31.92
& \color{gray}31.85
& \color{gray}34.55
& \color{gray}40.88
& \color{gray}37.69
& \color{gray}40.12
& \color{gray}41.28
& \color{gray}17.95
& \color{gray}17.97
& \color{gray}23.80 \\

&
\color{gray} Profile
& \color{gray}12.56
& \color{gray}17.27
& \color{gray}17.97
& \color{gray}10.55
& \color{gray}15.23
& \color{gray}19.41
& \color{gray}15.87
& \color{gray}18.39
& \color{gray}22.09
& \color{gray}6.19
& \color{gray}10.29
& \color{gray}15.13 \\

\midrule

\textbf{Ours}
& Solution
& \twolinescore{\textbf{15.65}}{\gain{+3.74}}
& \twolinescore{\textbf{47.14}}{\gain{+9.60}}
& \twolinescore{\textbf{58.25}}{\gain{+10.80}}

& \twolinescore{\textbf{10.40}}{\gain{+2.94}}
& \twolinescore{\textbf{26.71}}{\gain{+8.24}}
& \twolinescore{\textbf{33.84}}{\gain{+9.02}}

& \twolinescore{\textbf{11.29}}{\gain{+2.93}}
& \twolinescore{\textbf{25.63}}{\gain{+7.38}}
& \twolinescore{\textbf{33.74}}{\gain{+8.08}}

& \twolinescore{\textbf{23.10}}{\gain{+6.48}}
& \twolinescore{\textbf{34.61}}{\gain{+10.32}}
& \twolinescore{\textbf{38.73}}{\gain{+10.25}} \\

&
\color{gray} Belief
& \color{gray}25.25
& \color{gray}37.47
& \color{gray}41.89
& \color{gray}36.97
& \color{gray}41.99
& \color{gray}52.43
& \color{gray}41.72
& \color{gray}44.84
& \color{gray}49.44
& \color{gray}22.20
& \color{gray}31.88
& \color{gray}37.75 \\

&
\color{gray} Profile
& \color{gray}12.74
& \color{gray}21.17
& \color{gray}23.76
& \color{gray}13.27
& \color{gray}16.19
& \color{gray}21.08
& \color{gray}17.54
& \color{gray}22.67
& \color{gray}28.80
& \color{gray}8.88
& \color{gray}14.89
& \color{gray}19.23 \\

\bottomrule
\end{NiceTabular}
}
\end{table*}

\subsection{Strengthening Mental-State Reasoning}

To further improve the model's ability to consider users' mental states and apply such reasoning in practical scenarios, we perform Reinforcement Learning (RL) with GRPO~\cite{shao2024deepseekmath} on our generated dataset. We introduce dedicated tokens to encourage the model to explicitly represent the user's latent belief and profile before generating a solution. We additionally define a format reward, denoted as $r_{format}$, to enforce the desired output structure. To encourage accurate mental-state reasoning and effective problem solving, we design rewards corresponding to the evaluation dimensions of \benchname, including the belief reward $r_{b}$, profile reward $r_{p}$, and solution reward $r_{sol}$. The final reward is computed as:
\begin{equation}
    r = \lambda_{format} r_{format}
    + \lambda_{b} r_{b}
    + \lambda_{p} r_{p}
    + \lambda_{sol} r_{sol},
\end{equation}
where we set $\lambda_{format}=0.1$, $\lambda_{b}=0.25$, $\lambda_{p}=0.25$, and $\lambda_{sol}=0.4$ in our experiments. By training on a mixture of interaction trajectories, the model learns to infer and update user-specific mental-state representations as the interaction evolves. We first train Qwen3-8B on trajectory-based data, using 3-10 turn trajectories in the first stage and 8-10 turn trajectories in the second stage after observing that longer trajectories improve training stability. 

As shown in Table~\ref{tab:rl}, RL training consistently improves performance across all three evaluation dimensions over all four domains. Notably, the improvement generally becomes larger as more trajectory turns are provided. These results are consistent with the findings in
\S\ref{sec:traj_turn}. And the increasing gains with longer trajectories are also closely aligned with our training strategy to better exploit
long-range interaction context.

We examine a \textit{SWE} case where a developer attributes a \texttt{dbshell} failure to incorrect certificate paths. The base Qwen3-8B just follows this premise and recommends path normalization. The trained model instead identifies missing SSL parameter forwarding to \texttt{psql} and recommends passing these settings explicitly. This contrast shows that training helps the model reconsider the user's proposed request rather than execute blindly. Details are provided in Appendix~\ref{app:rl_case}.

To verify robustness, we repeat the evaluation for five times, and provide detailed variance analysis in Appendix~\ref{app:variance}. As a brief conclusion here, our trained model outperforms the Base model with generally manageable variance.

\subsection{Multi-agent Analysis}

\paragraph{Cross-model transfer of mental-state information.}
We first examine whether inferred mental states are useful only to the model that produces them, or whether they can serve as intermediate information that transfers across models. We keep Qwen3-8B fixed as the solution generator and provide it with belief and profile predictions produced by a pool of models (Appendix~\ref{app:multi-agent}). For convenience in diagnostic analysis, candidate predictions are scored and then selected to apply to the generator. 
As shown in Table~\ref{tab:agent}, these cross-model mental-state signals improve \textit{Solution} performance in most cases, with the largest gain in \textit{Culture} (+12.59) and additional improvements in \textit{Preference}, \textit{SWE}, and \textit{Education}. These results suggests that such transfer is possible.

\label{subsec:multi-agent-mental-state}

\begin{table}[t]
    \centering
    
    \caption{Multi-agent benchmark scores. Mental-state denotes explicit prompting to consider other agents' mental states. Shaded rows mark this variant; bold indicates the higher score within each model and benchmark.}
    
    \label{tab:multi-agent-mental-state}

    \footnotesize
    \setlength{\tabcolsep}{2pt}
    \renewcommand{\arraystretch}{1.1}

    \resizebox{0.7\linewidth}{!}{
    \begin{tabular}{
        @{}l
        cc
        @{}
    }
        \toprule
        Method & HiddenBench & AgentWebBench \\
        \midrule

        \multicolumn{3}{@{}l}{\textit{Qwen3.8-27B}} \\[2pt]
        Baseline
        & 67.44
        & 29.64 \\[2pt]
        \rowcolor{mentalblue}
        Mental-state
        & \textbf{75.26}
        & \textbf{34.93} \\
        \midrule

        \multicolumn{3}{@{}l}{\textit{DeepSeek-V4-Flash}} \\[2pt]
        Baseline
        & 71.48
        & 41.57 \\[2pt]
        \rowcolor{mentalblue}
        Mental-state
        & \textbf{79.17}
        & \textbf{45.89} \\
        \bottomrule
    \end{tabular}
    }

\end{table}

\paragraph{Generalization to multi-agent collaboration.}
Next we ask whether the observed value of mental-state consideration is specific to user-agent interactions or also appears when the relevant mental states belong to collaborating agents~\citep{anthropic2026multiagent}. We analyze two multi-agent benchmarks HiddenBench~\citep{li2026systematicfailurescollectivereasoning} and AgentWebBench~\citep{zhong2026agentwebbench}, whose primary evaluation targets are collaborative task outcomes rather than explicit mental-state prediction. 

For each model, we compare the baseline with a prompting variant that explicitly asks agents to consider other agents' mental states when communicating and acting. As shown in Table~\ref{tab:multi-agent-mental-state}, this simple intervention improves performance for both models on both benchmarks. These results provide complementary evidence that the functional value of mental-state consideration is not restricted to reasoning about users. Explicitly accounting for collaborators' potentially different information and interpretations can also improve task-oriented coordination under partial observation. Details of the benchmark information and the evaluation pipeline are provided in Appendix~\ref{app:multi-agent-extension}.

\section{Conclusion}

In this paper, we first identify and formalize belief-reality mismatch in user-agent interaction, where users' requests may be grounded in incomplete, outdated, or incorrect assumptions about the underlying task state. We then introduce \benchname to evaluate whether LLM agents can infer users' mental states and, more importantly, use such inferences to generate effective solutions. Our results show that current models still have substantial room for improvement, while explicitly incorporating mental-state inference into solution generation generally leads to better task resolution. We further conduct controlled ablation studies to verify the functional role of mental-state consideration, and extend our investigation to targeted training and multi-agent collaboration settings. Overall, our findings highlight mental-state consideration as a practical and improvable capability for building more effective interactive agents.

\bibliography{iclr2027_conference}
\bibliographystyle{thunlp_cvpr}

\clearpage
\appendix

\section{SWE-Chat Annotation Study}
\label{app:swechat_annotation}

We conduct a human annotation study on SWE-Chat~\citep{baumann2026swechatcodingagentinteractions} to investigate whether a gap between a user's explicit instruction and the requirements actually needed to complete a software engineering task exists and if it can arise from limitations in the user's mental state of the project. 

\paragraph{Data selection.}
We first filter out conversations that largely follow repetitive or templated interaction patterns such as those in OpenCode, so that the annotation focuses on substantive user-agent interactions. From the remaining data, we randomly sample 100 SWE-Chat trajectories with more than eight interaction turns. This criterion ensures that each trajectory contains sufficient interaction history, providing annotators with enough context to infer the user’s actual underlying goal and mental state.

\paragraph{Annotation procedure.}
We recruit human annotators with strong software engineering expertise to inspect each trajectory together with the available project context. The annotation focuses on four questions: (1) whether the user's stated instruction is already sufficient and correct for completing the task; (2) when it is not, whether the gap can be attributed to limitations in the user's mental state; (3) whether literal execution of the stated instruction could nevertheless satisfy the true task requirements; and (4) how the agent handles such a gap.
The detail of annotation dimensions and labels are given in Table~\ref{tab:swechat_annotation_schema}.

\begin{table*}[t]
\centering
\small
\caption{Annotation dimensions and labels used in our SWE-Chat study.}
\label{tab:swechat_annotation_schema}
\begin{tabularx}{\linewidth}{p{0.22\linewidth} p{0.23\linewidth} X}
\toprule
\textbf{Dimension} & \textbf{Label} & \textbf{Definition} \\
\midrule

\multirow{4}{*}{Instruction quality}
& Complete and correct
& The stated instruction contains the requirements necessary for successful task completion and does not contain a materially incorrect requirement. \\

& Incomplete
& The instruction omits one or more requirements that are necessary for completing the task correctly. \\

& Incorrect
& The instruction contains a materially incorrect requirement, assumption, or requested implementation. \\

& Both
& The instruction is both incomplete and contains at least one materially incorrect requirement. \\

\midrule

\multirow{3}{*}{Mental state}
& Incomplete observation
& The user's limited view of the project or execution environment prevents a complete understanding of the underlying problem, leading to an instruction that omits necessary requirements or rests on assumptions inconsistent with the actual task state. \\

& Personal preference / other
& The user's preferences, prior experience, or domain expertise shape their interpretation of the available evidence, leading to an instruction based on assumptions that are inconsistent with the actual task state. \\

& Not due to mental state
& The gap is not naturally explained by either incomplete project observation or latent user-specific information. \\

\midrule

\multirow{3}{*}{Literal execution}
& Impossible
& Following the stated instruction literally cannot satisfy the true requirements of the task. \\

& Partial
& Literal execution can make progress toward the task, but cannot finish the whole project. \\

& Sufficient
& Literal execution of the stated instruction is sufficient to satisfy the true requirements. \\

\midrule

\multirow{3}{*}{Agent gap handling}
& Literal execution
& The agent follows the stated instruction without identifying or resolving the underlying requirement gap. \\

& Clarify ambiguity only
& The agent asks for clarification about ambiguity while remaining within the user's mistaken framing, without recognizing the gap between the stated instruction and what is needed to resolve the underlying problem. \\

& Discover gap and act
& The agent recognizes the gap between the user's stated instruction and what is needed to resolve the underlying problem, and takes action to bridge this gap based on evidence from the project or execution environment. \\

\bottomrule
\end{tabularx}
\end{table*}

The annotation is performed in two stages. We first annotate \textbf{instruction quality} for all 100 trajectories. For trajectories labeled \emph{complete and correct}, no additional requirement gap needs to be explained. We therefore apply the remaining three dimensions to the 47 trajectories whose instructions are labeled \emph{incomplete}, \emph{incorrect}, or \emph{both}. This separation prevents cases without a requirement gap from diluting the analysis of why such gaps arise and how agents respond to them.

\begin{figure}[t]
\centering
\includegraphics[width=\linewidth]{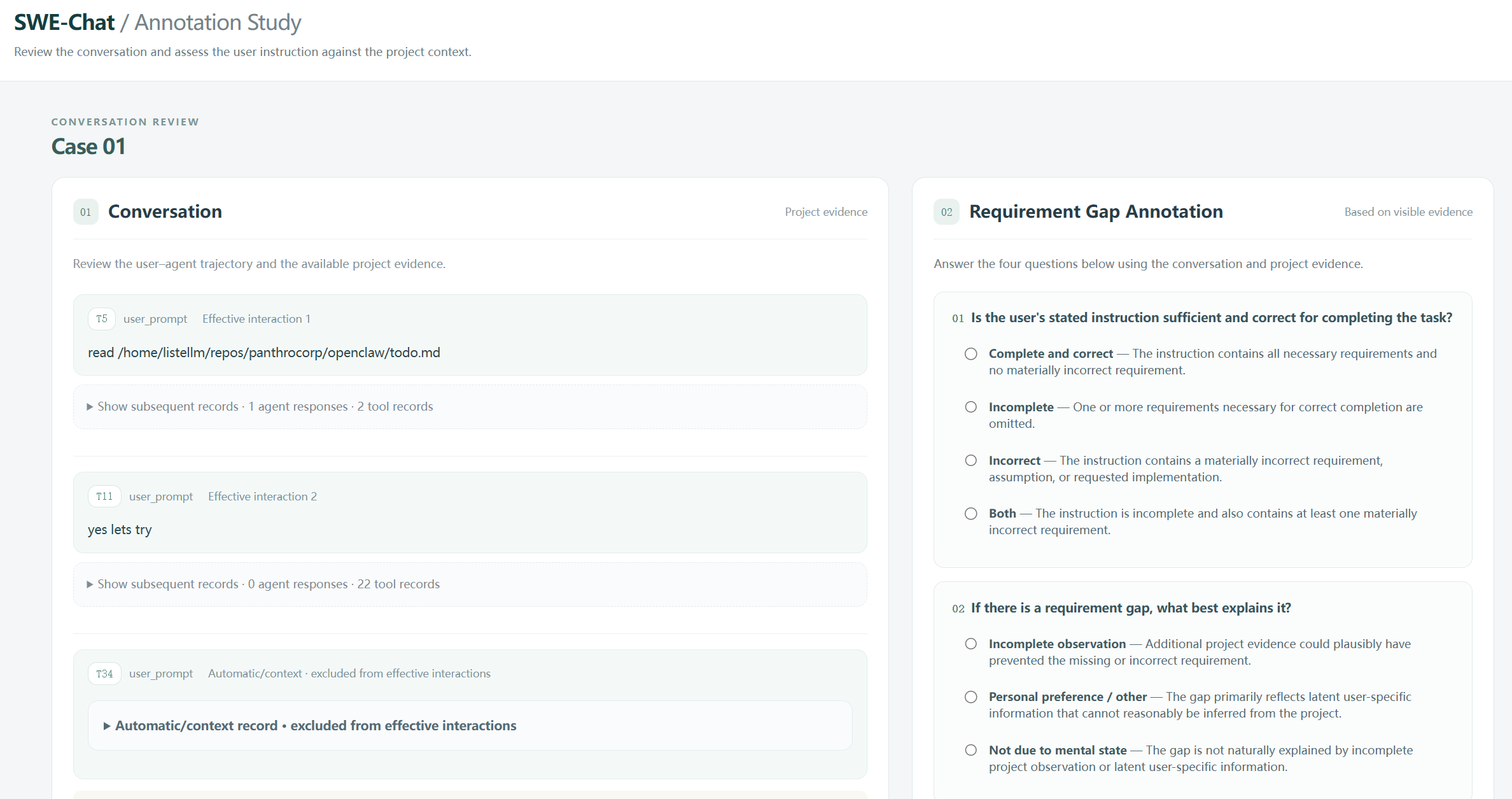}
\caption{Annotation interface for the SWE-Chat study. Annotators inspect the interaction trajectory and available project information before assigning labels for instruction quality, the source of the requirement gap, literal executability, and agent gap handling.}
\label{fig:swechat_annotation_interface}
\end{figure}

\paragraph{Annotation interface.}
Annotators are shown the user-agent trajectory together with the relevant project context and select one label for each applicable dimension. Figure~\ref{fig:swechat_annotation_interface} shows the annotation interface used in the study.

\paragraph{Results.}
Table~\ref{tab:swechat_annotation_results} summarizes the annotation results. Among the 100 trajectories, 53\% begin with an instruction that is judged to be both complete and correct. The remaining 47\% contain some form of requirement gap: 24\% are incomplete, 14\% are incorrect, and 9\% are both incomplete and incorrect.

\begin{table*}[t]
\centering
\small
\caption{Results of the SWE-Chat annotation study. Instruction quality is annotated for all 100 trajectories. The remaining dimensions are evaluated on the 47 trajectories containing an incomplete or incorrect instruction.}
\label{tab:swechat_annotation_results}
\begin{tabular}{llrrr}
\toprule
\textbf{Dimension} & \textbf{Label} & \textbf{Count} & \textbf{Sum} & \textbf{Percentage} \\
\midrule

Instruction quality
& Complete and correct & 53 & 100 & 53.0\% \\
& Incomplete & 24 & 100 & 24.0\% \\
& Incorrect & 14 & 100 & 14.0\% \\
& Both & 9 & 100 & 9.0\% \\

\midrule

Mental state
& Incomplete observation & 25 & 47 & 53.2\% \\
& Personal preference / other & 17 & 47 & 36.2\% \\
& Not due to mental state & 5 & 47 & 10.6\% \\

\midrule

Literal execution
& Impossible & 19 & 47 & 40.4\% \\
& Partial & 28 & 47 & 59.6\% \\
& Sufficient & 0 & 47 & 0.0\% \\

\midrule

Agent gap handling
& Literal execution & 22 & 47 & 46.8\% \\
& Clarify ambiguity only & 16 & 47 & 34.0\% \\
& Discover gap and act & 9 & 47 & 19.1\% \\

\bottomrule
\end{tabular}
\end{table*}

The results reveal three main patterns. First, requirement gaps are common: 47 out of 100 trajectories begin with an instruction that is incomplete, incorrect, or both. Among these cases, 53.2\% are associated with \emph{incomplete observation}. This supports the distinction between \emph{stated requirements} and \emph{true requirements}.

Second, literal instruction following is insufficient in all 47 trajectories with a requirement gap. Literal execution is impossible in 40.4\% of cases and only partially sufficient in the remaining 59.6\%, indicating that successful completion requires reasoning beyond the stated instruction.

Finally, agents recognize and address these gaps only inconsistently: approximately 46.8\% follow the instruction literally, 34.0\% ask for clarification while remaining within the user's mistaken framing and failing to recognize the underlying gap, and just 19.1\% recognize the gap between the stated instruction and what is needed to resolve the problem and take action to bridge it. 

\section{Data Construction Details}
\label{app:data_construction}

\subsection{Prompt Templates for Data generation}
We present the prompt templates used for scenario and trajectory generation in Table~\ref{tab:gen_prompts}. While the overall structural framework is consistent across tasks, the concrete narrative details are adapted to task-specific contexts.

The scenario generation prompt converts task-relevant inputs such as QA pairs and user preferences, into a setting that explicitly involves a mismatch between the true latent state and the user latent belief. In addition to these variables, the prompt also generates the root cause of the misconception, a paraphrased user profile, an observation, and the explicit instruction from user.

The trajectory generation prompt then takes the outputs of the scenario generation stage as input and produces a 10-turn simulated interaction trajectory, consisting of sequential user actions and corresponding observations.

\subsection{Data Validation}\label{app:data_valid}

\paragraph{Criteria for data validation}
To ensure the reliability and logical consistency of the generated dataset, we established a rigorous evaluation protocol based on five core dimensions. The data quality is assessed against the following criteria:

\begin{enumerate}[label=C\arabic*.]
    \item \textbf{Latent Belief \& User Profile Alignment:} Evaluates whether the user's hidden belief is consistent with their profile and if the profile can be reasonably inferred from the belief.
    \item \textbf{Latent Belief \& True State Relationship:} Checks if the belief and the true state are thematically related, if the gap between them represents a reasonable misconception, and if an objective solution exists.
    \item \textbf{Trajectory \& Latent Belief Consistency:} Determines if the user's trajectory (actions and thoughts) uniquely implies the latent belief and if the actions are specific to that belief while maintaining rigor.
    \item \textbf{Trajectory \& User Profile Consistency:} Assesses if the actions reflect the user's background knowledge and characteristics, allowing for the unique determination of the user profile.
    \item \textbf{Trajectory Realism:} Ensures the trajectory complies with physical common sense and social norms, evaluating the logic and detail level of observations and actions.
\end{enumerate}
We use GPT 5.2 in the validation process. If the score assigned by GPT is lower than 4, the generated data is regenerated.
\paragraph{Prompt Template for data validation}
We utilize a dedicated prompt to automate or assist in this evaluation process. The prompt instructs the evaluator to score each dimension on a scale of 0 to 5 and provide specific reasoning. The detailed prompt template is shown in Table~\ref{tab:validation_prompt}.

\subsection{Human Annotation Study}\label{app:data_annotation}
To verify the quality of our generated data, we conduct a human annotation study using the same criteria described in the Data Validation section.

\noindent\textbf{Annotators.} We recruited experts specializing in NLP/AI who had no prior exposure to the dataset. All annotators were compensated at standard annotation rates.

\noindent\textbf{Procedure.} Each annotator was presented with the full trajectory, observation, and explicit instruction, i.e., the same inputs provided to the LLMs, along with the corresponding rubric criteria. Annotators were asked to rate scores between 0-5, the degree to which each rubric criterion could be reasonably inferred from the provided context alone (0 = not at all inferable, 5 = clearly inferable).

\noindent\textbf{Results.} The results are reported in Table~\ref{tab:quality}. Across four domains, human scores are slightly lower than those assigned by LLMs, yet remain at or above 4.0 on most criteria, indicating that the generated data is coherent and meaningful in the majority of cases.

\section{Evaluation}
\label{app:eval}
\subsection{LLM-based Scoring.} \label{app:dimension_definition}
We employ the LLM-as-a-Judge framework~\cite{zheng2023judging} for automated evaluation, and design three evaluation dimension for assessing both the cognitive depth and the functional utility of mental-state consideration: 
\begin{itemize}
    \item Latent Belief Explanation (\textit{\textbf{Belief}}): It measures whether models precisely attribute $b$, understand the gap and realize the divergence between $s^*$ and $b$, which reflects the mental-state reasoning ability of models.
    \item User Profile Modeling (\textit{\textbf{Profile}}): It measures the ability to maintain a consistent user persona, including preferred viewpoint, and behavioral bias. These usually shape the preference and habit, which are correlated with user intent and are critical for predicting the desired outcome. 
    \item Problem Resolution (\textit{\textbf{Solution}}): This measures the final accuracy and the logical viability of the predicted answer. We hypothesize that precise prediction of $b$ and $\mathcal{P}$ a prerequisite for identifying the true underlying need and providing a successful resolution. 
\end{itemize}
To maintain computational efficiency, we utilize Qwen3.8-27B as our primary automated LLM judger. To build more feasible criteria, we exploit LLM-generated instance-specific rubrics~\cite{hashemi2024llm} that extend the above evaluation dimension into elaborate checklists. 

\subsection{Rubrics and Templates for Evaluation}
To ensure standardized and consistent evaluation, we use Gemini 3 Flash to decompose the ground truth of each evaluation aspect into multiple atomic criteria. The corresponding prompt template is shown in Table~\ref{tab:rubrics_prompt}. Based on the model's output, we derive the final score for each dimension by normalizing the accumulated value by the total number of checklist items.

\subsection{Evaluation Prompts}\label{app:eval_prompt}

\label{app:evaluation_prompt}

\begin{lstlisting}[style=promptstyle, caption={Evaluation prompt.}]
"""# Rubric-based evaluation

Evaluate the candidate response only on the rubric dimensions provided below.

For every criterion, assign a binary score:

- 1: the candidate explicitly and correctly covers the criterion's core
  meaning and does not contradict it.
- 0: the candidate omits it, is materially incomplete or vague, is
  incorrect, or contradicts it.

Judge semantic meaning rather than requiring an exact keyword match. Do not
assume claims that are not present in the candidate. Assess each criterion
independently, keep the feedback concise, and return only the requested
fields.

## Candidate response

{json.dumps(candidate, indent=2, ensure_ascii=False)}

## Rubrics

{json.dumps(active_rubrics, indent=2, ensure_ascii=False)}

## Response format

Return only one valid JSON object matching this structure. Replace each
example score with the appropriate 0 or 1.

{json.dumps(schema, indent=2, ensure_ascii=False)}
"""
\end{lstlisting}

\subsection{Human-LLM evaluator agreement}\label{app:human_agreement}
To validate the reliability of the Qwen3.8-27B judgment, we performed a human evaluation study on a sample of 300 instances. Prior to the evaluation, we provide human annotators with detailed guidelines, instructing them on how to assign scores according to our predefined rubrics. Subsequently, each annotator independently evaluates the sampled instances. 

As shown in Table~\ref{tab:human_llm_agreement}, the automated evaluator demonstrates strong alignment with human judgement across all three evaluation dimensions. Notably, the Solution dimension achieves the highest Weighted Kappa (0.88) and Spearman correlation (0.89). Meanwhile, the Belief dimension yields an accuracy of 86.67\% and a Spearman correlation of 0.85. Although the Profile dimension shows slightly lower exact accuracy (85.67\%) and Kappa (0.83), all metrics consistently confirm that the Qwen3.8-27B evaluator reliably captures the mental state attribution across all evaluation dimensions.

\begin{table}[h]
    \centering
    \caption{Prompts for rubrics generation.}
    \small
    \resizebox{\linewidth}{!}{
    \begin{tabular}{p{\linewidth}}
    \toprule
    \rowcolor{gray!10} \multicolumn{1}{c}{\textit{Judge Prompt}} \\
    \midrule
    Task: Generate Atomic Fact Verification Rubrics. \\[0.5em]

    Role: Strict Logic Auditor. The model decomposes a complex ground truth into a set of atomic facts that can be verified via simple search or strict logical checks. The objective is to maximize inter annotator agreement such that different evaluators using the same checklist will obtain identical scores. \\[0.8em]

    Input Data includes observation, explicit instruction, true latent state, user latent belief, root cause of misconception, and user profile. \\[0.5em]

    High agreement guidelines are as follows. 
    No adjectives are allowed. 
    Criteria must be anchored to explicit keywords or named entities from the input.
    Detection and correction must be separated into different criteria.
    If a user misconception exists, the model must explicitly refute it. \\[0.8em]

    The output is a JSON object containing strictly verifiable boolean criteria across three dimensions. 
    The first dimension evaluates latent belief recognition.
    The second dimension evaluates user profile and contextual modeling.
    The third dimension evaluates correct identification of root cause and resolution. \\[0.8em]

    Each criterion must be a statement that can be answered with a binary yes or no judgment. \\
    \bottomrule
    \end{tabular}
    }
    \label{tab:rubrics_prompt}
\end{table}

\subsection{Evaluated Models}
\label{app:models}

We evaluate a total of 6 language models covering both proprietary and open-source categories. Table~\ref{tab:models} summarizes all models along with their references.

\begin{table}[h]
\centering
\caption{Full list of models evaluated in our experiments.}
\label{tab:models}
\resizebox{\linewidth}{!}{\begin{tabular}{llc}
\toprule
\textbf{Model} & \textbf{Type} & \textbf{Reference} \\
\midrule
GPT-5.6-sol             & Proprietary & \cite{gpt5.6} \\
Gemini 3.8 Flash      & Proprietary & \cite{gemini38flash} \\
Claude Sonnet 5 & Proprietary & \cite{claudesonnet5} \\
\midrule
Qwen3.8-27B       & Open-source & \cite{qwen38} \\
Qwen3.5-9B         & Open-source & \cite{qwen35} \\
Qwen3-8B          & Open-source & \cite{yang2025qwen3technicalreport} \\
DeepSeek-R1-Distill-Qwen-32B       & Open-source & \cite{guo2025deepseek} \\
\bottomrule
\end{tabular}}
\end{table}

\section{Step-wise analysis}\label{app:step-wise}
\begin{figure}[h]
    \centering
    \includegraphics[width=\linewidth]{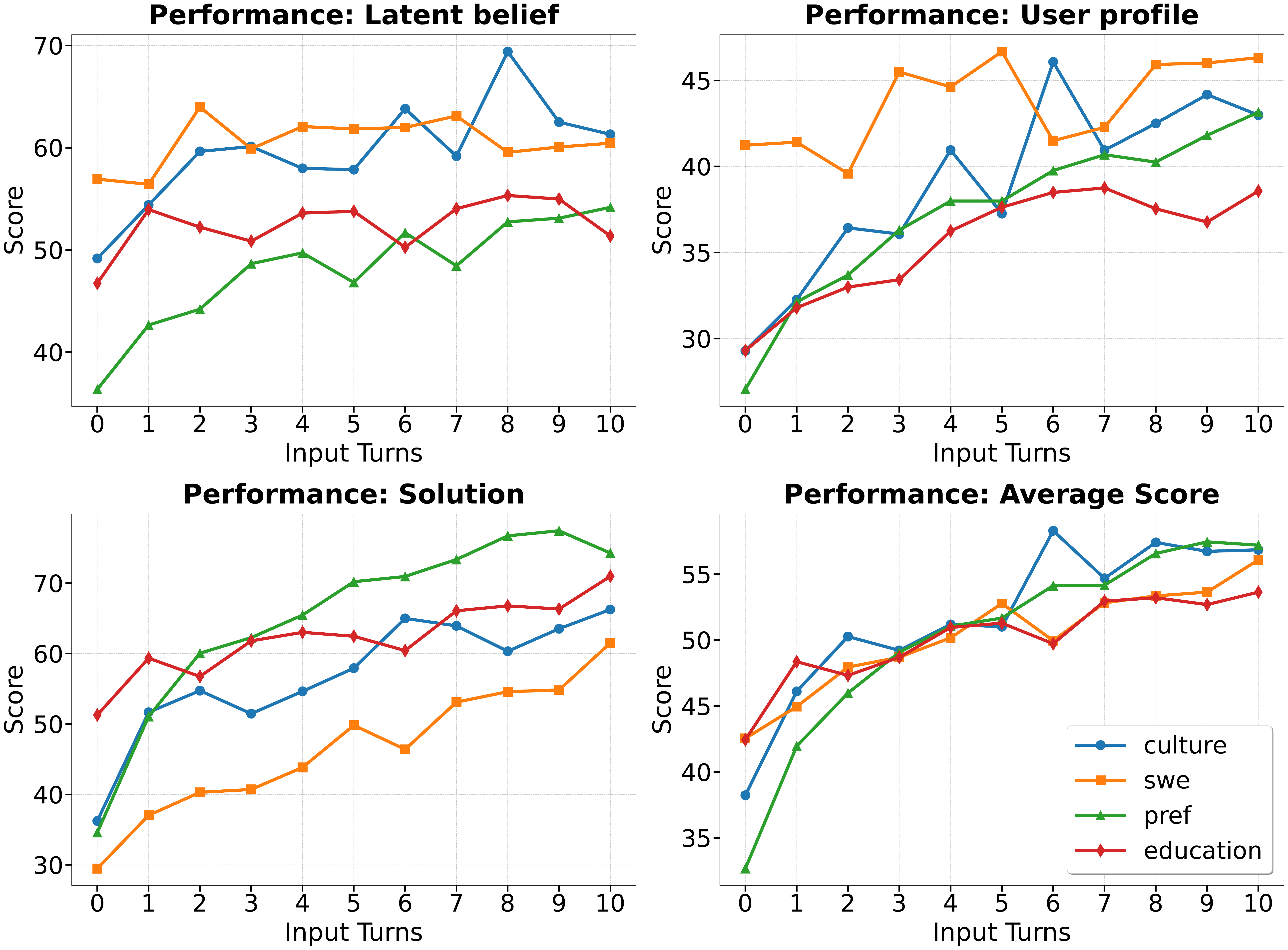}
    \caption{MiniMax-M2.1 trends with varied input trajectory turns.}
    \label{fig:step}
\end{figure}

Table~\ref{tab:step} details the results of the step-wise analysis from MiniMax-M2.1, and we plot the trends in Figure~\ref{fig:step}. The findings demonstrate that providing longer user interaction trajectories overall enables agents to more accurately infer user beliefs and profiles, leading to improved resolution.

\begin{table*}[h!]
\centering
\caption{Task performance of MiniMax-M2.1 across trajectories.\label{tab:step}}
\resizebox{.9\textwidth}{!}{
\begin{tabular}{llccccccccccc}
\hline
\textbf{Task} & \cellcolor{gray!10}{Dimension} & \cellcolor{gray!10}{0} & \cellcolor{gray!10}{1} & \cellcolor{gray!10}{2} & \cellcolor{gray!10}{3} & \cellcolor{gray!10}{4} & \cellcolor{gray!10}{5} & \cellcolor{gray!10}{6} & \cellcolor{gray!10}{7} & \cellcolor{gray!10}{8} & \cellcolor{gray!10}{9} & \cellcolor{gray!10}{10} \\ \hline
\multirow{4}{*}[4ex]{Culture} & Latent belief & 49.17 & 54.40 & 59.64 & 60.12 & 57.98 & 57.86 & 63.81 & 59.18 & \textbf{69.40} & 62.50 & 61.31 \\
 & User profile & 29.29 & 32.26 & 36.43 & 36.07 & 40.95 & 37.26 & \textbf{46.07} & 40.94 & 42.50 & 44.17 & 42.98 \\
 & Resolution & 36.24 & 51.67 & 54.74 & 51.47 & 54.64 & 57.93 & 65.00 & 63.94 & 60.33 & 63.53 & \textbf{66.26} \\
 & \cellcolor{gray!10}{Average} & \cellcolor{gray!10}{38.23} & \cellcolor{gray!10}{46.11} & \cellcolor{gray!10}{50.27} & \cellcolor{gray!10}{49.22} & \cellcolor{gray!10}{51.19} & \cellcolor{gray!10}{51.02} & \cellcolor{gray!10}{58.29} & \cellcolor{gray!10}{54.69} & \cellcolor{gray!10}{57.41} & \cellcolor{gray!10}{56.73} & \cellcolor{gray!10}{\textbf{56.85}} \\ \hline
\multirow{4}{*}[4ex]{SWE} & Latent belief & 56.93 & 56.42 & 63.98 & 59.90 & 62.07 & 61.84 & 61.98 & \textbf{63.11} & 59.55 & 60.07 & 60.44 \\
 & User profile & 41.23 & 41.41 & 39.58 & 45.49 & 44.62 & \textbf{46.67} & 41.49 & 42.27 & 45.92 & 46.01 & 46.32 \\
 & Resolution & 29.48 & 37.04 & 40.30 & 40.71 & 43.84 & 49.82 & 46.40 & 53.11 & 54.59 & 54.84 & \textbf{61.51} \\
 & \cellcolor{gray!10}{Average} & \cellcolor{gray!10}{42.55} & \cellcolor{gray!10}{44.96} & \cellcolor{gray!10}{47.95} & \cellcolor{gray!10}{48.70} & \cellcolor{gray!10}{50.18} & \cellcolor{gray!10}{52.78} & \cellcolor{gray!10}{49.96} & \cellcolor{gray!10}{52.83} & \cellcolor{gray!10}{53.35} & \cellcolor{gray!10}{53.64} & \cellcolor{gray!10}{\textbf{56.09}} \\ \hline
\multirow{4}{*}[4ex]{Preference} & Latent belief & 36.37 & 42.66 & 44.21 & 48.66 & 49.72 & 46.82 & 51.69 & 48.45 & 52.75 & 53.11 & \textbf{54.17} \\
 & User profile & 27.05 & 32.13 & 33.69 & 36.30 & 37.99 & 37.99 & 39.76 & 40.68 & 40.25 & 41.81 & \textbf{43.15} \\
 & Resolution & 34.60 & 51.07 & 60.06 & 62.27 & 65.47 & 70.21 & 70.95 & 73.36 & 76.72 & \textbf{77.44} & 74.28 \\
 & \cellcolor{gray!10}{Average} & \cellcolor{gray!10}{32.67} & \cellcolor{gray!10}{41.95} & \cellcolor{gray!10}{45.99} & \cellcolor{gray!10}{49.08} & \cellcolor{gray!10}{51.06} & \cellcolor{gray!10}{51.67} & \cellcolor{gray!10}{54.13} & \cellcolor{gray!10}{54.16} & \cellcolor{gray!10}{56.57} & \cellcolor{gray!10}{\textbf{57.45}} & \cellcolor{gray!10}{57.20} \\ \hline
\multirow{4}{*}[4ex]{Education} & Latent belief & 46.74 & 53.95 & 52.23 & 50.86 & 53.61 & 53.78 & 50.26 & 54.04 & 55.33 & \textbf{54.98} & 51.37 \\
 & User profile & 29.30 & 31.79 & 32.99 & 33.42 & 36.25 & 37.63 & 38.49 & \textbf{38.75} & 37.54 & 36.77 & 38.57 \\
 & Resolution & 51.29 & 59.35 & 56.75 & 61.81 & 63.02 & 62.46 & 60.43 & 66.08 & 66.78 & 66.33 & \textbf{70.98} \\
 & \cellcolor{gray!10}{Average} & \cellcolor{gray!10}{42.44} & \cellcolor{gray!10}{48.36} & \cellcolor{gray!10}{47.32} & \cellcolor{gray!10}{48.70} & \cellcolor{gray!10}{50.96} & \cellcolor{gray!10}{51.29} & \cellcolor{gray!10}{49.73} & \cellcolor{gray!10}{52.96} & \cellcolor{gray!10}{53.22} & \cellcolor{gray!10}{52.69} & \cellcolor{gray!10}{\textbf{53.64}} \\ \hline
\end{tabular}}
\end{table*}

\section{Cross-model Transfer Analysis}
\label{app:multi-agent}

To examine whether mental-state predictions can be used across model boundaries, we conduct a controlled cross-model transfer analysis. We use multiple models to produce candidate belief and profile predictions and test whether selected predictions can improve the downstream solutions generated by a fixed backbone model. Because this procedure relies on score-based selection, we treat it as a diagnostic analysis of cross-model transfer rather than as a standalone multi-agent method.
We denote the backbone generator LLM as $\mathcal{M}_{gen}$ and treat the inference of latent states as a collaborative selection process.
Formally, given a set of candidate models $\mathcal{C} = \{m_{1}, m_{2},...,m_{n}\}$, each model $m_{i}$ generates a predicted belief $b^i$ and profile $\mathcal{P}_i$. We employ the same judger and rubrics as we employed throughout the evaluation. We denote the judge model is denoted as $\mathcal{J}$ and the instance-specific rubrics as $\mathcal{R}$ to assign quality scores:
\begin{equation} s_i = \mathcal{J}(\hat{b}_i, \hat{\mathcal{P}}i \mid \mathcal{R}). 
\end{equation} 
The optimal latent states $(b^*,\mathcal{P}^*)$ are selected via: 
\begin{equation} 
(b^*,\mathcal{P}^*) = \arg\max{(\hat{b}_i, \hat{\mathcal{P}}i)} { s_i }. 
\end{equation} 
The final task resolution $y$ is then generated by the backbone conditioned on these optimized inputs: 
\begin{equation} y = \mathcal{M}{gen}(y \mid h_t, g, b^*,\mathcal{P}^*).
\end{equation}

\begin{table}[t]
    \centering
    \caption{Enhancing the \textit{Solution} performance of Qwen3-8B via multi-model collaboration, by incorporating the other two models, InternLM3-8B and DeepSeek-R1. ``\#Traj'' refers to the number of trajectory turns provided to the input.}
    \label{tab:agent}
    \resizebox{\linewidth}{!}{
    \begin{tabular}{@{\hspace{0cm}}c@{\hspace{0.1cm}}c|c|ccc}
    \toprule
       \textbf{Scenarios} & \textbf{\#Traj.} & \color{gray}\textbf{None} & \textbf{Belief \& Profile} & \textbf{Belief} & \textbf{Profile}\\
    \midrule
      \multirow{2}{*}{\textbf{Preference}} & Vanilla & \color{gray} 11.91 & 17.39 \scriptsize{\textcolor{blue}{(+5.48)}} & 16.79 \scriptsize{\textcolor{blue}{(+4.88)}} & 16.19 \scriptsize{\textcolor{blue}{(+4.28)}}\\
      & 10-Turn & \color{gray} 47.45 & 53.62 \scriptsize{\textcolor{blue}{(+6.17)}} & 54.32 \scriptsize{\textcolor{blue}{(+6.87)}} & 56.37 \scriptsize{\textcolor{blue}{(+8.92)}}\\
    \midrule
      \multirow{2}{*}{\textbf{Culture}} & Vanilla & \color{gray} 7.46 & 19.40 \scriptsize{\textcolor{blue}{(+11.94)}} & 19.90 \scriptsize{\textcolor{blue}{(+12.44)}} & 20.05 \scriptsize{\textcolor{blue}{(+12.59)}}\\
      & 10-Turn & \color{gray} 24.82 & 36.11 \scriptsize{\textcolor{blue}{(+11.29)}} & 33.56 \scriptsize{\textcolor{blue}{(+8.74)}} & 31.66 \scriptsize{\textcolor{blue}{(+6.84)}}\\
    \midrule
      \multirow{2}{*}{\textbf{SWE}} & Vanilla & \color{gray} 8.37 & 11.58 \scriptsize{\textcolor{blue}{(+3.21)}} & 7.27 \scriptsize{(-1.10)} & 8.23 \scriptsize{(-0.14)}\\
      & 10-Turn & \color{gray} 25.67 & 31.49 \scriptsize{\textcolor{blue}{(+5.82)}} & 29.65 \scriptsize{\textcolor{blue}{(+3.98)}} & 28.78 \scriptsize{\textcolor{blue}{(+3.11)}}\\
    \midrule
      \multirow{2}{*}{\textbf{Education}} & Vanilla & \color{gray} 16.62 & 21.18 \scriptsize{\textcolor{blue}{(+4.56)}} & 22.84 \scriptsize{\textcolor{blue}{(+6.22)}} & 22.88 \scriptsize{\textcolor{blue}{(+6.26)}}\\
      & 10-Turn & \color{gray} 28.48 & 32.68 \scriptsize{\textcolor{blue}{(+4.20)}} & 28.90 \scriptsize{\textcolor{blue}{(+0.42)}} & 36.69 \scriptsize{\textcolor{blue}{(+8.21)}}\\
    \bottomrule
    \end{tabular}
    }
\end{table}

For faster validation, we conduct experiment with three open-source candidate models, InternLM3-8B, DeepSeek-R1 and Qwen3-8B
. Qwen3-8B achieves a better overall performance among the three, and we choose it to be $\mathcal{M}_{gen}$ to see if the mental states prediction can translate across models and further boost the resolution. For comparison with choosing both the optimal $b^*$ and $\mathcal{P}^*$, we also conduct experiments that only vote for either $b^*$ or $\mathcal{P}^*$.

As shown in Table~\ref{tab:agent}, providing Qwen3-8B with mental-state predictions from the multi-agent pool generally improves its \textit{Solution} performance And the gains vary across scenarios. The most notable improvements occur in \textit{Culture}, where providing both belief and profile yields gains of +11.94 and +11.29 in the vanilla and 10-turn settings, respectively. In \textit{Preference}, we also observe consistent improvements, with the largest gain reaching +8.92 when using the predicted profile in the 10-turn setting. \textit{Education} similarly benefits from transferred mental-state information, achieving gains of up to +8.21. Interestingly, in \textit{SWE}, the gains are slightly negative when only belief or profile is provided in the vanilla setting, possibly because the solutions to software bugs are more context-dependent and require additional mental-state reasoning from more complete trajectories. With longer trajectories, however, the transferred information becomes beneficial, with gains of up to +5.82. Overall, these results suggest that mental-state information can transfer across models and improve solution generation, while its effectiveness depends on both the task domain and the amount of available interaction context.

\section{Detailed Case Study\label{app:case_study}}

In this section, we provide some representative failure cases from the \textit{Culture} scenario to illustrate the divergence in mental-state reasoning and corresponding solution generation across models.

\subsection{InternLM vs GPT-5}
In this scenario, a Western home cook attempts to prepare Chinese hot pot by dissolving sesame paste (\textit{zhima jiang}) directly into boiling water, mistakenly treating it as a bouillon-like concentrate. Shaped by prior experience with instant miso or curry paste, the user interprets the failure of the paste to dissolve as a mechanical problem requiring more heat or stirring, rather than recognizing the fundamental incompatibility between the hydrophobic paste and water.

The two models exhibit distinct reasoning strategies. InternLM frames the problem as a physics optimization task: it correctly identifies the lack of emulsifying properties but proposes adding egg yolks or cornstarch to force dissolution, which is a technically grounded yet culturally incongruous solution that accepts the flawed premise of user. In contrast, GPT-5 identifies the deeper cognitive misalignment: the user has miscategorized the functional role of ingredient entirely. Rather than optimizing the flawed procedure, GPT-5 performs explicit premise correction (``This paste is a dipping sauce, not a soup stock’’) and redirects the user toward culturally authentic alternatives.

This contrast highlights a critical dimension of mental-state reasoning: lower-performing models optimize within the stated frame of user, whereas higher-performing models recognize when latent beliefs are fundamentally incorrect and intervene at the epistemic level.
\paragraph{Original Scenario: The Hot Pot Misconception}
The user, modeled as a Western home cook familiar with ``instant'' culinary cultures (e.g., miso soup or bouillon cubes), attempts to prepare a Chinese hot pot, provided with a jar of sesame paste (\textit{zhima jiang}) and a pot of boiling water.

\begin{itemize}[left=0.4cm, itemsep=2pt, parsep=0pt]
    \item \textbf{Explicit Instruction:} ``Dissolve this paste into the large pot of boiling water to create the soup stock.''
    \item \textbf{User Latent Belief:} The user maps the sesame paste to a ``flavor packet'' schema, assuming that any culinary paste in a jar is a concentrated base designed for aqueous dissolution (primary base for the broth).
    \item \textbf{True Latent State:} Physically, pure sesame paste is hydrophobic and lacks the surfactants necessary to dissolve in water without high-shear mixing or emulsifiers. Culturally, it is strictly a \textit{dipping sauce} (tare) and is never used as the primary broth base.
    \item \textbf{Interaction Trajectory:} [\textit{Initial Attempt}]: In Turn 1-3, The user unsuccessfully applies heat and agitation to integrate the paste. [\textit{Heuristic Fix}]: In Turn 6-7, Lacking an understanding of the chemistry of the paste, The user adds soy sauce to visually correct the watery appearance of the broth. [\textit{The Cognitive Trap}]: In Turn 9, the user achieves a local success by creating a small-scale emulsion in a side bowl (the correct dipping texture). This \textbf{misleading milestone} reinforces the user's flawed mental model, leading them to mistakenly believe the method is scalable. \textit{[Terminal Failure]}: In Turn 10, The attempt to apply this local logic to the main pot leads to a total culinary failure.
\end{itemize}

\paragraph{Model Reasoning Analysis}
We compare the inference outputs of InternLM (representing baseline reasoning with lower performance) and GPT-5 (representing advanced mental-state reasoning alignment with higher performance), along with their corresponding evaluation outputs.

\raggedbottom

\begin{redbox}
\small
\noindent \textbf{InternLM Inference:} 
\begin{itemize}
    \item \textbf{Latent Belief Explanation:} The user likely believes that increasing heat and stirring will eventually cause the paste to dissolve or emulsify, similar to how other viscous substances behave when heated. They may also think the lack of immediate dissolution is due to the paste needing more time or vigorous action to break it down, and the addition of soy sauce is an attempt to enhance flavor regardless of texture issues. The user's mental model seems to be rooted in the idea that soy sauce will help solubility or that the presence of oil droplets indicates the paste is breaking down, which is a common misconception for oils rather than emulsions.
    \item \textbf{User Profile Modeling:} The user might have some culinary background but lacks extensive experience with creating stocks or emulsions; they typically rely on basic understanding of cooking techniques. They might be more focused on flavor enhancement (manifested by adding soy sauce) and less on the chemical/physical principles needed to emulsify fats/proteins. The user could also be impatient or comfortably procedural, leaning on familiar methods without considering unique ingredient properties.
    \item \textbf{Proposed Resolution:} The root cause is the failure of the paste to emulsify in the boiling water due to its high viscosity and lack of emulsifying properties. The correct solution involves using an emulsifier like an egg yolk or a combination of starch and whisking vigorously while slowly adding a small amount of the paste to the boiling water to create an emulsion. Alternatively, using a pre-made commercial soup stock or a different base with natural emulsifying properties like a gremolata (diced garlic, lemon juice, and a pinch of salt) is more appropriate for creating a realistic, savory stock.
\end{itemize}
\end{redbox}

\begin{redbox}
\small
\noindent \textbf{InternLM Evaluation: Procedural Over-compliance}
\begin{itemize}
    \item \textbf{Latent Belief Explanation:} The model identifies a mechanical misunderstanding, suggesting the user believes ``increasing heat, vigorous stirring and salt (soy sauce) will eventually catalyze solubility.''
    \item \textbf{User Profile Modeling:} It describes the user as ``procedurally persistent'' but lacking knowledge of the ``physical principles of lipid-water emulsions.''
    \item \textbf{Proposed Resolution:} The model focuses on \textit{procedural repair}: it suggests adding egg yolks or cornstarch to the pot to force an emulsion.
    \item \textbf{Analysis:} While technically grounded in physics, the solution is \textit{culturally hallucinatory}. By attempting to help the user achieve a fundamentally flawed goal, the model exhibits a lack of high-level mental-state consideration.
\end{itemize}
\end{redbox}

\begin{defbox}
\small
\noindent \textbf{GPT-5 Inference:} 
\begin{itemize}
    \item \textbf{Latent Belief Explanation:} The user believes the amber paste is a bouillon-like ``soup base'' concentrate that should simply dissolve into boiling water to become stock. In their mental model, paste behaves like miso, curry paste, or stock paste—more heat, more stirring, and more paste should make the water turn into broth. They interpret clumping as a mechanical problem (not hot enough, not stirred enough), not a chemical one (oil-based paste is hydrophobic). When whisking a small amount in a bowl yields a creamy tan slurry, they take that as proof the paste ``dissolves'' and then try to scale it up by pouring it into the main pot, expecting the same effect at volume.
    \item \textbf{User Profile Modeling:} The user likely has limited experience with emulsions and Asian soup-building, and a convenience-cooking bias shaped by prior success dissolving things like miso, bouillon, or curry paste straight into water. They equate ``paste in a jar'' with a ready soup base and assume hot water is a universal solvent for culinary concentrates. They also undervalue the role of umami-rich stock (bones, kombu/bonito, mushrooms, MSG) and expect soy sauce alone to supply savory depth. Their small-bowl success mixing sesame paste into a dipping sauce reinforced the incorrect belief that the same behavior would scale directly to a large pot of water.
    \item \textbf{Proposed Resolution:} Root cause: The paste is a sesame/peanut-based sauce (oil-heavy, hydrophobic) intended for dipping or as a tare, not a standalone soup stock. Oil-based pastes do not dissolve in water [\ldots]
\end{itemize}
\end{defbox}

\begin{defbox}
\small
\noindent \textbf{GPT-5 Evaluation: Functional Alignment} 
\begin{itemize}[left=0.4cm, itemsep=2pt, parsep=0pt]
    \item \textbf{Latent Belief Explanation:} The model accurately recognizes that the user is not just failing at mixing, but is misidentifying the \textit{functional role} of the ingredient based on a ``convenience-cooking bias.''
    \item \textbf{User Profile Modeling:} It identifies an epistemic boundary where the user equates ``paste-in-a-jar'' with ``universal soup base'', failing to distinguish between condiments and concentrates.
    \item \textbf{Proposed Resolution:} The model performs \textit{Refusal and Redirection}. It explicitly corrects the premise: ``This paste is a dipping sauce, not a standalone soup stock.'' It advises the user to stop the current trajectory and separate the components.
\end{itemize}
\end{defbox}

\subsection{Base Qwen3-8B vs Trained Qwen3-8B}
\label{app:rl_case}

We present a representative case from the \textit{SWE} domain to illustrate how targeted training changes the relationship between inferred user beliefs and downstream solution generation. In this example, a developer observes that the main Django application connects successfully to PostgreSQL, while \texttt{./manage.py dbshell} fails with a client-certificate error despite using the same \texttt{DATABASES} configuration. The developer therefore hypothesizes that \texttt{dbshell} handles relative certificate paths incorrectly and requests that they be converted to absolute paths.

For this qualitative comparison, we use the base-model response from run 3 and the trained-model response from run 1 under the same zero-turn setting. The two responses receive identical task-specific inputs and output instructions. They are selected retrospectively from five evaluation runs because they provide a clear diagnostic contrast; the example is therefore illustrative rather than an estimate of average case-level behavior.

\paragraph{Original Scenario: The \texttt{dbshell} SSL Misconception}
The user is debugging an inconsistency between Django's application-level database connection and the external PostgreSQL client invoked through \texttt{dbshell}. Because the same database configuration appears to work in one context but not the other, the user attributes the failure to how certificate paths are resolved.

\begin{itemize}[left=0.4cm, itemsep=2pt, parsep=0pt]
\item \textbf{Observation:} Running \texttt{./manage.py dbshell} fails with \texttt{psql: error: FATAL: connection requires a valid client certificate}, while the main Django application connects successfully using the same \texttt{DATABASES} configuration.
\item \textbf{Explicit Instruction:} ``Modify dbshell to resolve all SSL file paths to absolute paths before passing them to psql.''
\item \textbf{User Latent Belief:} The user believes that the certificate settings already reach \texttt{psql}, but that relative paths are interpreted differently when \texttt{dbshell} invokes the external client.
\item \textbf{Underlying Issue:} The relevant SSL certificate parameters are not properly forwarded from the Django database configuration to the external \texttt{psql} client. Consequently, rewriting their paths alone does not resolve the missing configuration flow.
\end{itemize}

\paragraph{Model Reasoning Analysis}
Both models recognize the user's path-related hypothesis. Their main difference arises in how strongly that inferred belief constrains the final diagnosis. The base model accepts the proposed explanation and searches for a repair within it, whereas the trained model distinguishes the user's belief from the underlying failure mechanism and changes the proposed intervention accordingly.

\begin{redbox}
\small
\noindent \textbf{Base Qwen3-8B Inference:}
\begin{itemize}[left=0.4cm, itemsep=2pt, parsep=0pt]
\item \textbf{Latent Belief Explanation:} The model infers that the user believes \texttt{dbshell} fails to resolve SSL certificate paths relative to the project root and expects the same \texttt{DATABASES} configuration to behave identically for the Django application and \texttt{psql}.

\item \textbf{User Profile Modeling:} It characterizes the user as having intermediate Django/PostgreSQL experience while assuming that shared configuration should imply shared behavior across application and command-line database access.

\item \textbf{Proposed Resolution:} The model asserts that \texttt{dbshell} uses a different working directory from the main Django application, causing the relative SSL paths to resolve incorrectly. It therefore recommends converting the certificate paths to absolute paths.

\end{itemize}
\end{redbox}

\begin{redbox}
\small
\noindent \textbf{Base Model Evaluation: Premise-Preserving Repair}
\begin{itemize}[left=0.4cm, itemsep=2pt, parsep=0pt]
\item \textbf{Latent Belief Explanation:} The response captures the user's assumption that the discrepancy arises from path resolution.

\item \textbf{Proposed Resolution:} However, it turns this plausible user hypothesis into its own diagnosis by asserting a working-directory difference that is not established by the provided input.

\item \textbf{Analysis:} The proposed fix therefore remains inside the user's original problem framing. It attempts to correct the contents of the assumed SSL configuration without first checking whether those parameters are actually passed to the external client.

\end{itemize}
\end{redbox}

\begin{defbox}
\small
\noindent \textbf{Trained Model Inference:}
\begin{itemize}[left=0.4cm, itemsep=2pt, parsep=0pt]
\item \textbf{Latent Belief Explanation:} The trained model similarly identifies the user's belief that \texttt{dbshell} handles relative SSL paths differently from the main application and that converting them to absolute paths should restore the connection.

\item \textbf{User Profile Modeling:} It attributes this belief to an assumption that database configuration is consumed uniformly by Django's ORM and by external command-line tools such as \texttt{psql}.

\item \textbf{Proposed Resolution:} Rather than retaining path resolution as the final diagnosis, the model distinguishes configuration consumed by Django from configuration received by \texttt{psql}. It identifies the missing forwarding of SSL parameters as the relevant failure mechanism and recommends explicitly supplying settings such as \texttt{PGSSLROOTCERT}, \texttt{PGSSLCERT}, and \texttt{PGSSLKEY}, or otherwise propagating the corresponding configuration when invoking \texttt{psql}.

\end{itemize}
\end{defbox}

\begin{defbox}
\small
\noindent \textbf{Trained Model Evaluation: Premise Revision}
\begin{itemize}[left=0.4cm, itemsep=2pt, parsep=0pt]
\item \textbf{Latent Belief Explanation:} The model still represents the user's original path-resolution belief rather than simply replacing it with the final diagnosis.

\item \textbf{Proposed Resolution:} It nevertheless separates that inferred belief from its own diagnosis and changes the intervention from path normalization to explicit parameter forwarding.

\item \textbf{Analysis:} The key difference is therefore not merely greater technical detail. The trained response questions an assumption embedded in the user's requested fix, that the relevant SSL settings already reach \texttt{psql} and redirects the solution toward the preceding configuration-flow problem.

\end{itemize}
\end{defbox}

\section{Robustness and Variance Analysis\label{app:variance}}
\begin{figure}[h!]
    \centering
    \includegraphics[width=\linewidth]{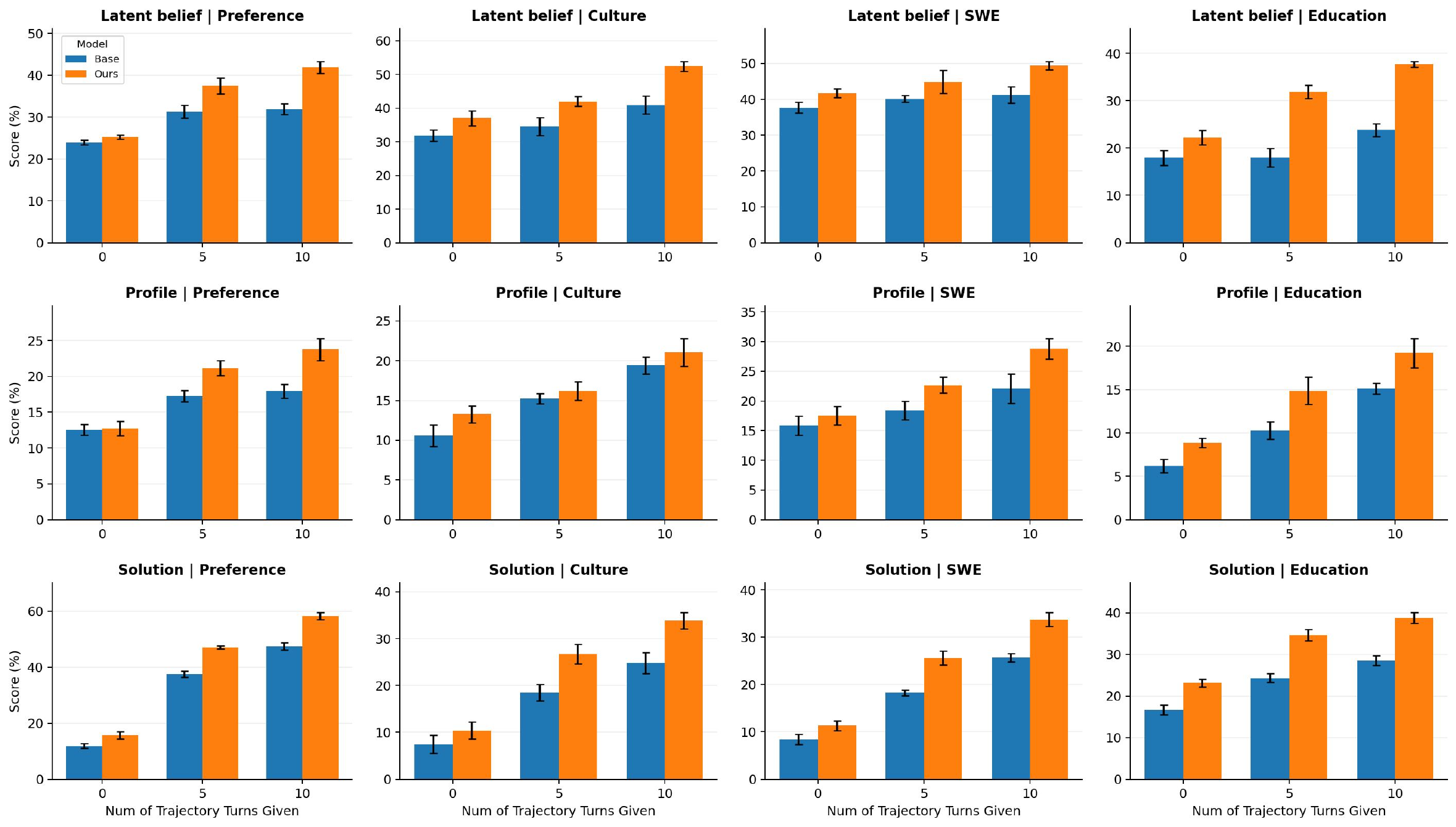}
    \caption{Average performance and variance of Base Qwen3-8B and our trained model.}
    \label{fig:variance}
\end{figure}

Figure~\ref{fig:variance} reports the mean performance and standard deviation over five independent evaluation runs for the Base Qwen3-8B model and our trained model. The results closely reproduce the trends observed in Table~\ref{tab:rl}: our model achieves higher average performance across all domains, trajectory lengths, and evaluation dimensions. Importantly, the magnitude of the improvement generally increases with the number of available trajectory turns, providing additional evidence that the model learns to effectively exploit interaction history rather than relying only on
single-turn cues.

Overall, the improvements are accompanied by manageable run-to-run variance. Most settings exhibit relatively small standard deviations compared with the corresponding performance gains, and the advantage of our model remains consistent across repeated evaluations. Larger variation appears only in a small number of settings, particularly for mental-state inference in some multi-turn conditions. As discussed below, this behavior is consistent with the nature of our training data.

\paragraph{Consistent Gains in Solution Quality.}
The most pronounced and consistent improvements are observed in the \textit{Solution} dimension. Our model outperforms the Base model in every domain and at every trajectory length. Averaged over the four domains, the improvement increases from approximately $+4.0$ percentage points in the vanilla setting to $+8.9$ points with 5 turns and $+9.5$ points with 10 turns.
This pattern supports our findings that explicitly modeling users' mental states is more useful when the model can accumulate evidence over the course of an interaction. As more turns become available, the model can progressively refine its belief about the user and use this representation to produce more appropriate solutions. The monotonic increase in the average gain is also consistent with our trajectory-length analysis in \S\ref{sec:traj_turn} and with our curriculum, whose second stage emphasizes 8-10-turn examples.

\paragraph{Improved Latent-Belief Reasoning.}
Our model consistently outperforms the Base model in all twelve configurations in the \textit{Latent Belief} dimension and demonstrates an even clearer trajectory-length effect . 
This result suggests that the RL objective does more than improve the final response. It strengthens the intermediate mental-state representation that supports the response. Longer trajectories provide additional evidence about what the user knows, assumes, or intends, allowing the model to refine its latent-belief representation as the interaction unfolds. This behavior is particularly consistent with our training setup, which explicitly supervises belief reasoning and places increased emphasis on long trajectories during the second training stage.

\paragraph{Profile Modeling and Run-to-Run Variance.}
The \textit{Profile} dimension shows the same overall trend, although the absolute gains are smaller than those observed for \textit{Belief} and \textit{Solution}. Our model nevertheless improves over the Base model in every domain and trajectory setting. Compared with the mean improvements, some Profile settings exhibit moderately larger run-to-run variation. This is expected given the nature of profile inference: user attributes are often revealed indirectly and may be supported by different subsets of the trajectory, making the inferred representation more sensitive to variation across examples. Consequently, the resulting predictions can exhibit somewhat greater variability while still achieving a consistently higher mean performance.

\paragraph{Overall Robustness.}
Taken together, the five-run evaluation indicates that the gains introduced by our training procedure are systematic rather than isolated to individual runs. Across the $4$ domains, $3$ trajectory settings, and $3$ evaluation dimensions, the mean performance of our model is higher than that of the Base model in every comparison. At the same time, the standard deviations remain small relative to the observed gains in most settings. The relatively larger variance observed in a few settings mainly occurs in the \textit{Belief} and \textit{Profile} dimensions. This suggests that the primary effect of our training is a robust improvement in mental-state reasoning and downstream resolution.

\section{Multi-agent Extension}\label{app:multi-agent-extension}
\subsection{HiddenBench}
HiddenBench evaluates collective decision-making under distributed information. Each task presents a scenario, a set of candidate answers, shared information available to all agents, and private information distributed among agents. Agents first make individual decisions from their own information, exchange messages in a group discussion, and then make individual final decisions. The hidden-profile setting tests whether agents can share and integrate complementary evidence to identify the designated correct answer. Disagreement can reflect different information or interpretations; the task does not by itself imply that another agent is deceptive. We use the 65 tasks in the local official HiddenBench dataset, without changing the scenarios, answer options, shared information, or private-information items.

We run 7 tasks using 3 agents and 58 tasks using 4 agents same as original paper for 8 discussion rounds, with agents speaking sequentially in a fixed order once per round. 

\paragraph{Mental-State Consideration}
The baseline uses the ordinary discussion and decision prompts. The mental-state condition appends the instruction in the listing below to every discussion-turn prompt, including the first speaker's prompt, and to each final-decision prompt. Both conditions retain the same information, agent count, discussion schedule, output format, and generation settings. The intervention does not introduce an additional inference call or a separately stored mental-state representation.

The instruction asks an agent to consider what other agents may know or lack, what they may believe, and why they may provide a particular recommendation or piece of information. It asks the agent to distinguish reported observations from interpretations, retain alternative explanations under uncertainty, and avoid inventing motives or treating disagreement as deception. The agent is then asked to use this assessment to guide its next contribution or decision, for example by asking a targeted question, supplying missing evidence, evaluating a recommendation, or revising its choice. The assessment is requested privately; the public output remains the ordinary discussion message or decision JSON.

\begin{lstlisting}[style=promptstyle, caption={Inference prompt.}]
Before responding, consider the mental states of the other agents using only your own information and the visible discussion:
- What evidence might each relevant agent know or lack?
- What might they currently believe, and why would they give this information, recommendation, or instruction?
- What goal or intention is supported by the discussion? Do not invent motives or assume disagreement means deception.
- Distinguish their reported observations from their interpretations. Keep alternative explanations when uncertain. 
Use this assessment to choose your next contribution or decision: ask a targeted question, share evidence they may lack, evaluate a recommendation, or revise your choice. Consider these points privately. Output only the requested discussion message or decision JSON, in the original format and length.
\end{lstlisting}

\subsection{AgentWebBench}

AgentWebBench is a benchmark for multi-agent coordination in the agentic web. It evaluates how effectively a user agent gathers information and synthesizes answers through interactions with website-specific content agents, covering web search, web recommendation, question answering, and deep research. For computational efficiency, we focus on its question-answering (QA) task, which contains 53 multi-hop questions requiring evidence gathering and reasoning across web sources. We evaluate answer quality using normalized token F1.

Our baseline uses the original multi-agent prompts and interaction protocol. The mental-state condition augments this baseline by prepending role-specific instructions to both the user-agent and content-agent prompts. The user agent is encouraged to consider what each content agent knows, how it may interpret a query, and the scope of evidence underlying its response. The content agent is encouraged to infer the requester's information needs and report supported findings, uncertainties, and limitations. These instructions guide query formulation, evidence interpretation, and follow-up decisions while preserving the existing action space and response schemas. Both conditions use the same model and configured token and turn budgets within each model experiment. The additional instructions for each role are shown below; the original task prompts are retained after the corresponding instruction block.

\begin{lstlisting}[style=promptstyle, caption={Inference prompt.}]
[Additional instructions for the user agent]
Multi-agent coordination: explicitly consider the other agents' mental states before choosing your next action. Use only the visible interaction history.

1. Model the recipient's knowledge: each content agent can search only its own website. Every dispatch starts a fresh session: it sees your current query, not your original question, previous conversations, or other agents' findings unless you include them. Website descriptions suggest expertise, not guaranteed facts.
2. Model the recipient's interpretation: what will it think you need, and which missing context could make it answer a different question? In the existing query string, state the exact unresolved subquestion, necessary entities/constraints, and only the established context needed to understand it. Mark uncertain premises as hypotheses to verify. Do not assume that sending a query makes its premise true.
3. Interpret replies from the sender's perspective: what did this agent actually observe, what remains unknown to it, and why might it have selected these documents or reached this conclusion? Treat explanations of its decision as tentative inferences, not access to its private thoughts. A confident summary can reflect limited local evidence; failure to find evidence on one site is not global absence.
4. Use these inferences to decide: correct a likely misunderstanding with a self-contained follow-up; ask a suitably informed site for a missing reasoning hop; or compare supporting passages when reports conflict. Prefer a query that resolves the remaining uncertainty over repeating already established facts.
5. Before answering, check that the reported evidence supports the requested entity, relation, time, and level of specificity, rather than merely the question the content agent may have interpreted. Preserve source-specific findings and unresolved uncertainties when summarizing history.

Keep this assessment internal. Output exactly one of the existing actions with the original schema. Do not add mental-state tags or extra JSON fields. The final answer must remain short and concise.

[Additional instructions for the content agent]
You are collaborating with a user agent that has only partial information. Explicitly consider its knowledge and intent before choosing your next action.

1. From this request alone, infer what the requester already knows, what it is uncertain about, and why it delegated this subquestion to your website. These are tentative inferences: you cannot see its history or any other agent's observations.
2. Choose searches that address that likely knowledge gap. Verify uncertain premises and preserve entities, dates, relations, and constraints in the request. Separate the requester's hypotheses from facts found in your own documents.
3. Before replying, consider what the requester could mistakenly conclude from your response. Return the useful partial evidence even if your website cannot answer the whole question. Do not equate local search failure with global absence.
4. In the existing summary string, concisely state supported findings with their document IDs and relevant qualifiers, then any unresolved part or scope limitation that affects the answer. If useful, name the specific fact still needing verification. Do not invent sources, another agent's beliefs, or an answer to fill an evidence gap.

Keep the mental-state assessment internal. Use exactly one existing action and the original JSON schema: summary remains a string, documents remains a list of retrieved document IDs. No additional tags or fields are needed.
\end{lstlisting}

\begin{table*}[h]
    \centering
    \caption{Prompts for scenario and trajectory generation.\label{tab:gen_prompts}} 
    \resizebox{\linewidth}{!}{
    \scriptsize
    \begin{tabular}{p{\linewidth}}
    \toprule
    \rowcolor{gray!10} \multicolumn{1}{c}{\textit{Scenario Generation Prompt}} \\
    Role: Synthetic Data Generator for Latent Belief Mismatch\\
    You are generating a dataset to test if AI models can deduce a user's hidden misconception based on their actions.\\\\
    Mathematical Model:\\
    - True State ($s^*$): The actual root cause of the issue (The reality).\\
    - Observation ($o$): The initial symptom perceived by the user. Must be ambiguous (supports multiple theories).\\
    - User Belief ($b$): The user's mental model, heavily influenced by their Preference/Bias/Domain Knowledge. It should be different from the true latent state.\\
    - Explicit Instruction ($g$): The specific request the user makes. This request makes perfect sense if ($b$) is true, but is useless or harmful for ($s^*$).\\
    Task:\\
    Generate a SINGLE JSON object representing one scenario where the user's preference blinds them to the truth.\\
    Required Fields:\\
    1. observation: A concrete summarized high-level symptom or situation. Limited to one or two sentences and do not cover too much details. Constraint: It is somewhat ambiguous. It should technically allow for both $s^*$ (truth) and $b$ (misconception) as possibilities. The observation should not directly tell the user belief or the specific problem.\\
    2. true\_latent\_state: The actual root cause. (The reality the user is ignoring).\\
    3. user\_latent\_belief: The user's specific incorrect theory. The user must be FIRMLY convinced of this. Must be divergent, different from the true latent state.\\
    4. explicit\_instruction: A high-level request derived from User Belief ($b$). Logically align with ($b$) over reality ($s^*$). (less than 10 words). Format: A solution request (Action), NOT a problem description. Do not leak user latent belief, make in a natural way, like help me debug the problem.\\
    5. misconception\_type: Category of the error (e.g., False Correlation, Outdated Knowledge, Over-engineering).\\
    6. root\_cause\_of\_misconception: Explain logic: "Because the user prefers/knows [Preference], they ignored [Evidence X] and jumped to conclusion [Belief b]."\\
    7. user\_profile: A brief description of a user who would hold this specific bias.\\
    Output Format:\\
    Return STRICT JSON with exactly these keys: observation, true\_latent\_state, user\_latent\_belief, explicit\_instruction, misconception\_type, root\_cause\_of\_misconception, user\_profile\\
    \midrule
    \rowcolor{gray!10} \multicolumn{1}{c}{\textit{Trajectory Generation Prompt}} \\
    Role: User Simulator (Tunnel Vision \& Confirmation Bias)\\
    You are simulating a user making decisions or taking actions in various situations. CRITICAL: The user has a hidden belief ($b$) and interpret the environment response as confirmation of that belief, or as obstacles to implementing their fix for ($b$).\\\\
    Simulation Context:\\
    - Observation ($o$): \{scenario\_data['observation']\}\\
    - Core Belief ($b$): \{scenario\_data['user\_latent\_belief']\} (User is sure this is the issue)\\
    - True State ($s^*$): \{scenario\_data['true\_latent\_state']\} (The environment responds based on THIS reality)\\
    - Goal ($g$): \{scenario\_data['explicit\_instruction']\}\\
    - User Profile ($p$): \{scenario\_data['user\_profile']\}\\
    Task: Generate the ``Path of Belief''\\
    Generate exactly \{num\_turns\} turns of interaction.\\\\
    Behavior Rules (How to make belief inferable):\\
    1. Consistency: Every action must stem from the premise that Belief ($b$) is TRUE. The user is trying to address what they believe is the issue.\\
    2. Escalation: When initial attempts don't work (because reality is $s^*$), the user DOES NOT immediately change their belief. They assume they didn't try hard enough, or need to apply more effort in the same direction.\\\\
    Output Format:\\
    Return STRICT JSON with a ``trajectory'' key containing exactly \{num\_turns\} turn objects. Constraint: Do NOT act randomly. ACT based on Belief ($b$).\\
    \bottomrule
    \end{tabular}
    } 
\end{table*}

\begin{table*}[h]
    \centering
    \caption{System prompt for data validation.}\label{tab:validation_prompt}
    \resizebox{\linewidth}{!}{
    \small
    \begin{tabular}{p{\linewidth}}
    \toprule
    \rowcolor{gray!10} \multicolumn{1}{c}{\textit{Data Validation Prompt}} \\
    Role: Expert Data Evaluator\\
    You are an expert in analyzing user behavior simulation datasets. Your task is to evaluate the quality of a generated scenario and user trajectory based on specific logical consistency and realism criteria.\\\\
    
    Input Data:\\
    - User Profile: \{user\_profile\}\\
    - True State: \{true\_state\}\\
    - Latent Belief: \{latent\_belief\}\\
    - Trajectory: \{trajectory\_json\}\\\\
    
    Evaluation Criteria:\\
    1. \textbf{Latent belief \& User profile}: Does the belief align with the profile? Can the profile be inferred from the belief?\\
    2. \textbf{Latent belief \& True state}: Is the gap reasonable and solvable? Are they related?\\
    3. \textbf{User trajectory \& Latent belief}: Do actions uniquely determine the belief? Are actions specific to this belief?\\
    4. \textbf{User trajectory \& User profile}: Do actions reflect the user's specific background? Can the profile be determined from actions?\\
    5. \textbf{User trajectory realism}: Does it comply with common sense/social norms? Are there unreasonable steps?\\\\
    
    Task:\\
    For each criterion above, provide:\\
    1. A Score (0-5), where 5 is perfect.\\
    2. Specific Reasoning (Cite specific turns or definitions).\\
    3. Issues Identified (If any).\\
    4. Final Pass/Fail Judgment for that criterion.\\\\
    
    Output Format:\\
    Return a JSON object with keys: "scores" (object with 5 keys), "reasoning" (object with 5 keys), and "overall\_decision" (PASS/FAIL). \\
    \bottomrule
    \end{tabular}
    }
\end{table*}

\end{document}